\documentclass[letterpaper,conference]{IEEEtran}
\usepackage[pdftex]{graphicx} \pdfcompresslevel=9
\usepackage{subfig}
\usepackage{stmaryrd}
\usepackage{amsfonts}

\usepackage[all,frame,poly,arc]{xy}

\newtheorem{thm}{Theorem}[section]

\IEEEoverridecommandlockouts    

\begin{document}

\title{A Growing Self-Organizing Network for Reconstructing Curves and
Surfaces\thanks{The author is with the Laboratorio di Visione
Artificiale, Universit\`a degli	Studi di Pavia, Via Ferrata, 1 - 27100 Pavia,
Italy (email: marco.piastra@unipv.it).}}

\author{Marco Piastra}

\maketitle
\thispagestyle{empty}

\begin{abstract}
   Self-organizing networks such as Neural Gas, Growing Neural Gas and
   many others have been adopted in actual applications for both dimensionality
   reduction and manifold learning. Typically, in these applications, the
   structure of the adapted network yields a good estimate of the topology of
   the unknown subspace from where the input data points are sampled. The
   approach presented here takes a different perspective, namely by assuming
   that the input space is a manifold of known dimension. In return, the new
   growing self-organizing network introduced here gains the ability to adapt
   itself in way that may guarantee the effective and stable recovery of the
   exact topological structure of the input manifold.
\end{abstract}

\section{Introduction}

In the original Self-Organizing Map (SOM) algorithm by Teuvo Kohonen  
\cite{Kohonen97} a lattice of connected units learns a representation of an
input data distribution. During the learning process, the weight vector -
i.e. a position in the input space - associated to each unit is progressively
adapted to the input distribution by finding the unit that best matches each
input and moving it `closer' to that input, together with a subset of
neighboring units, to an extent that decreases with the distance on the
lattice from the best matching unit. As the adaptation progresses, the SOM tends
to represent the topology input data distribution in the sense that it maps
inputs that are `close' in the input space to units that are neighbors in the
lattice.

In the Neural Gas (NG) algorithm \cite{Martinetz-etal91}, the topology of the
network of units is not fixed, as it is with SOMs, but is learnt from the input
distribution as part of the adaptation process. In particular, Martinetz and
Schulten have shown in \cite{Martinetz-Schulten94} that, under certain
conditions, the Neural Gas algorithm tends to constructing a \emph{restricted
Delaunay graph}, namely a triangulation with remarkable topological properties
to be discussed later. They deem the structure constructed by the algorithm a
\emph{topology representing network} (TRN).

Besides the thread of subsequent developments in the field of neural networks,
the work by Martintetz and Schulten have raised also a considerable interest
in the community of computational topology and geometry. The studies that followed in 
this direction have produced a number of theoretical results that are nowadays
at the foundations of some popular methods for curve and surface reconstruction
in computer graphics (\cite{Cazals-Giesen06}), although they have little or
nothing in common with neural networks algorithms.

Perhaps the most relevant of these results, for the purposes of what follows,
is the assessment of the theoretical possibility to reconstruct from a point
sample a structure which is homeomorphic to the manifold that coincides with
the support of the sampling distribution (see definitions below).
Homeomorphism, as we will see, is a stronger condition than being just
restricted Delaunay and highly desirable, too. The main requirement for
achieving this is ensuring the proper density of the `units' in the structure
with respect to the features of the input manifold.

\begin{figure}[t] 
  \centering
  \includegraphics[width=.7\linewidth]{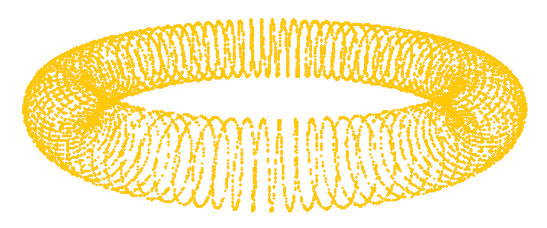}
  \includegraphics[width=.7\linewidth]{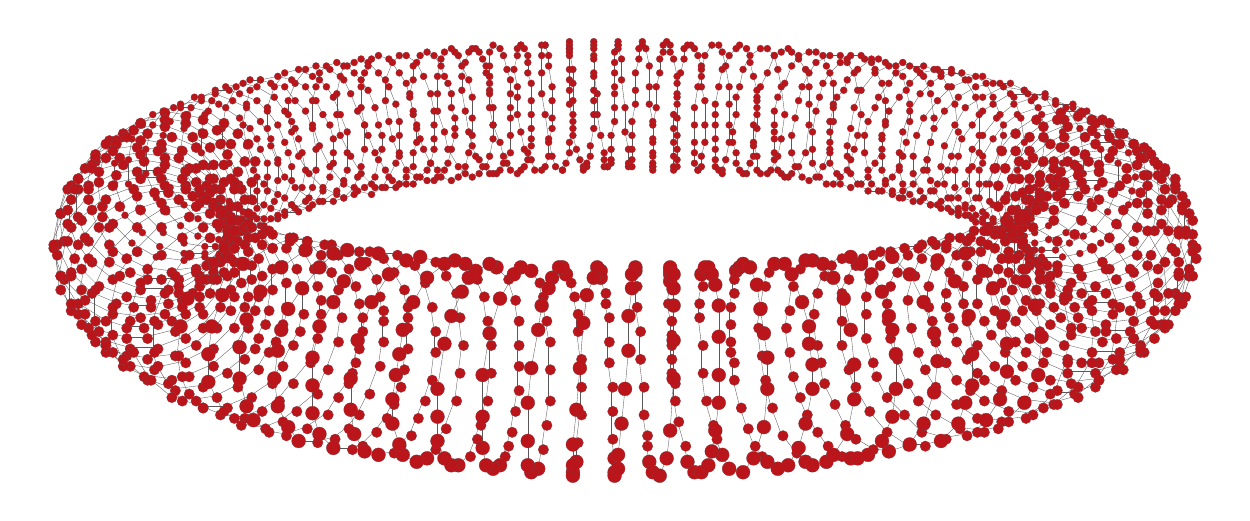}
  \includegraphics[width=.7\linewidth]{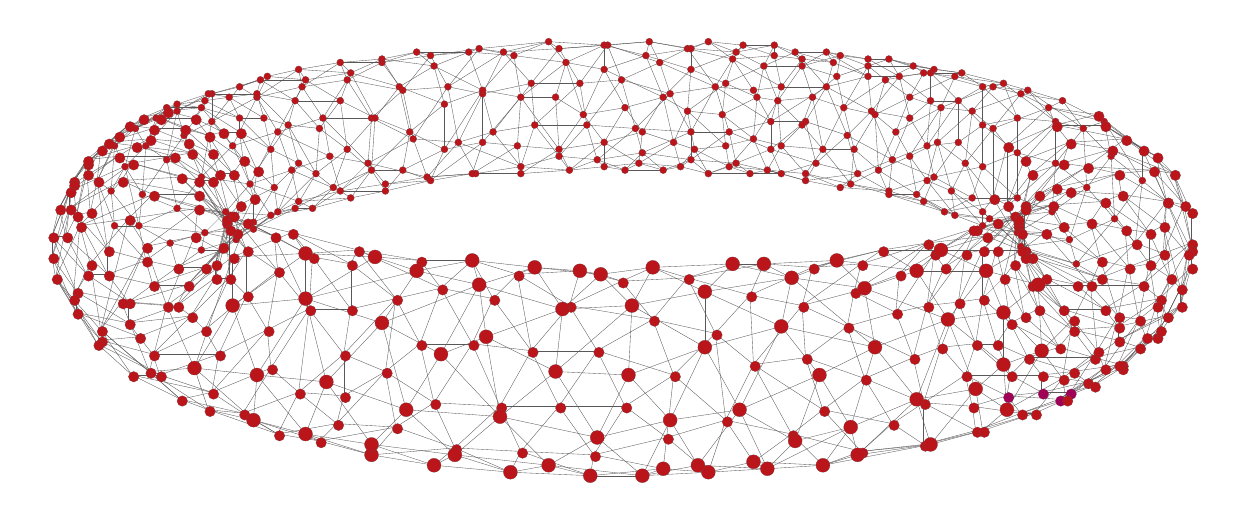}
  \caption{\label{fig:htorus}Depending on its internal settings, from a point
  sample of an helical curve running on a torus, a SOAM will either reconstruct
  a curve (in the middle) or a surface (below).}
\end{figure}
The main contribution of this work is showing that the above theoretical result
can be harnessed in the design of a new kind of growing neural network that
automatically adapts the density of units until, under certain condition to
be analyzed in detail, the structure becomes homeomorphic to the input
manifold. Experimental evidence shows that the new algorithm is effective with
a large class of inputs and suitable for practical applications.

\section{Related work}

The Neural Gas algorithm \cite{Martinetz-etal91} also introduces the so-called
\emph{competitive Hebbian rule} as the basic method for establishing connections
among units: for each input, a connection is added, if not already present,
between closest and second-closest unit, according to the metric of choice. In
order to cope with the mobility of units during the leaning process, an aging
mechanism is also introduced: at each input, the age of the connection between
the closest and second-closest units, if already present, is refreshed, while
the age of other connections is increased by one. Connections whose age
exceeds a given threshold will eventually be removed.

As proven in \cite{Martinetz-etal93}, the NG algorithm obeys a stochastic
gradient descent on a function based on the average of the geometric
quantization error. As known, this is not true of a SOM \cite{ErwinEtAl92},
whose learning dynamics does not minimize an objective function of any sort.
This property of NG relies on how the units in the network are adapted: at
each input, the units in the network are first sorted according to their
distance and then adapted by an amount that depends on the ranking.

A well-known development of the NG algorithm is the Growing Neural Gas (GNG)
\cite{Fritzke95}. In GNG, as the name suggests, the set of units may grow (and
shrink) during the adaptation process. Each unit in a GNG is associated to a
variable that stores an average value of the geometric quantization error.
Then, at fixed intervals, the unit with the largest average error is detected
and a new unit is created between the unit itself and the neighbor unit having
the second-largest average error. In this way, the set of units grows
progressively until a maximum threshold is met. Units in GNG can also be
removed, when they remain unconnected as a result of connection aging.
In the GNG algorithm the adaptation of positions is limited to the immediate
connected neighbors of the unit being closest to the input signal. This is
meant to avoid the expensive sorting operation required in the NG algorithm is
to establish the ranking of each unit. By this, the time complexity $O(N \log
N)$ of each NG iteration can be reduced to linear time in the number of units
$N$. This saving comes at a price, however, as the convergence assessment of
NG described in \cite{Martinetz-etal93} does not apply anymore. The GNG
algorithm is almost identical to the Dynamic Cell Structure by Bruske and
Sommers \cite{Bruske-Sommer95}.

In the Grow-When-Required (GWR) algorithm \cite{Marsland-etal02}, which is a
development of GNG, the error variable associated to each GNG unit is replaced
by a firing counter $f$ that decreases exponentially each time the unit is
winner, i.e. closest to the input signal. When $f$ gets below a certain
threshold $T_{f}$, the unit is deemed \emph{habituated} and its behavior
changes. The adaptation of \emph{habituated} unit being closest to the input
takes place only if the distance is below a certain threshold $R$, otherwise a
new unit is created. This means that the network continues to grow until the
input data sample is completely included in the union of balls of radius $R$
centered in each unit.

\section{Topological interlude}
The basic definitions given in this section will be necessarily quite concise.
Further information can be found in textbooks such as \cite{Edelsbrunner06}
and \cite{Zomorodian05}.

Two (topological) spaces $X$ and $Y$ are said to be \emph{homeomorphic}, denoted
as $X \approx Y$, if there is a function $f : X \rightarrow Y$ that is
bijective, continuous and has a continuous inverse. A (sub)space $M \subseteq
\mathbb{R}^d$ is a closed, compact \emph{k-manifold} if every point $x \in M$
has a \emph{neighborhood} (i.e. an open set including it) that is
\emph{homeomorphic} to $\mathbb{R}^k$. In other words, a $k$-manifold is a
(sub)space that locally `behaves' like $\mathbb{R}^k$.

\begin{figure}[tcb]
  \centering
  \subfloat[]{\includegraphics[width=.47\linewidth]{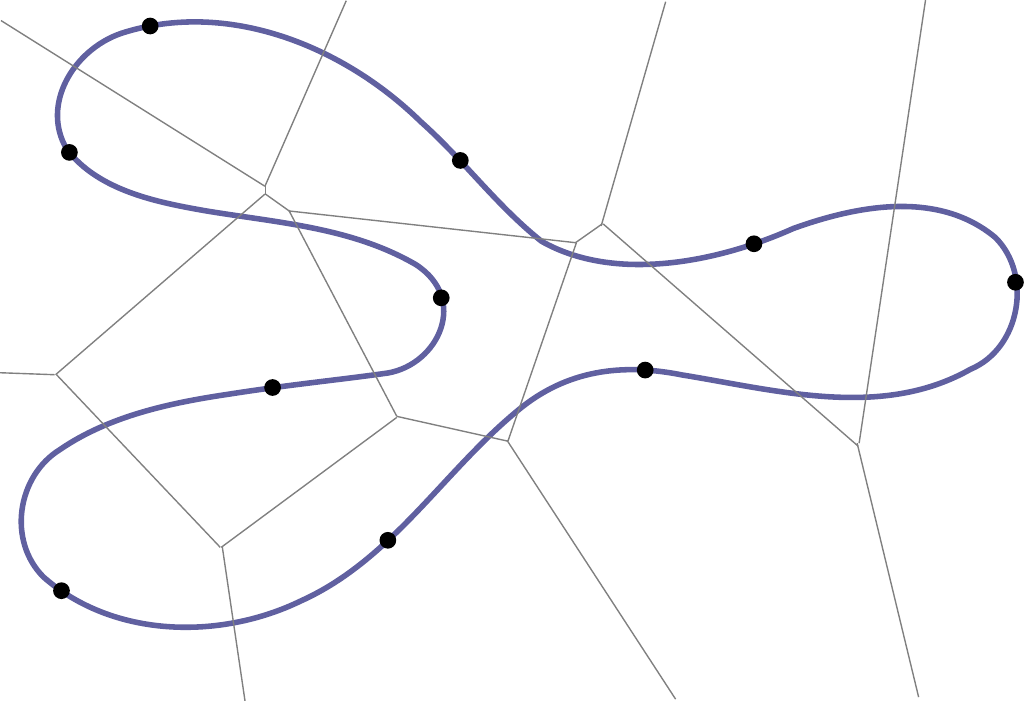}}
  \hspace{.01\linewidth}
  \subfloat[]{\includegraphics[width=.47\linewidth]{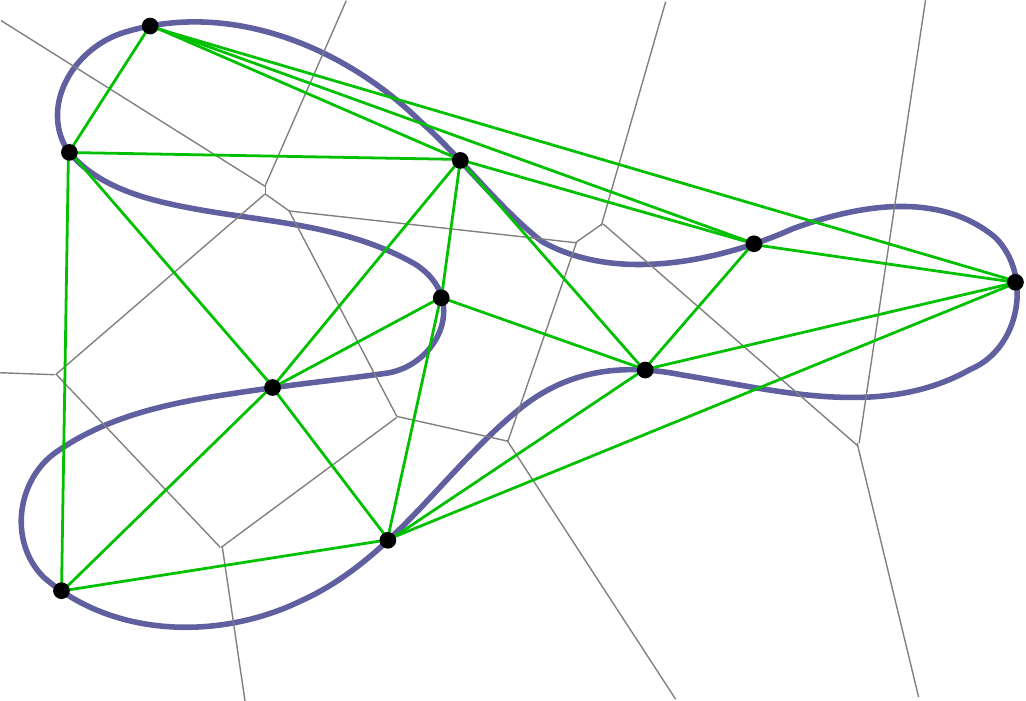}}
  \\
  \subfloat[]{\includegraphics[width=.47\linewidth]{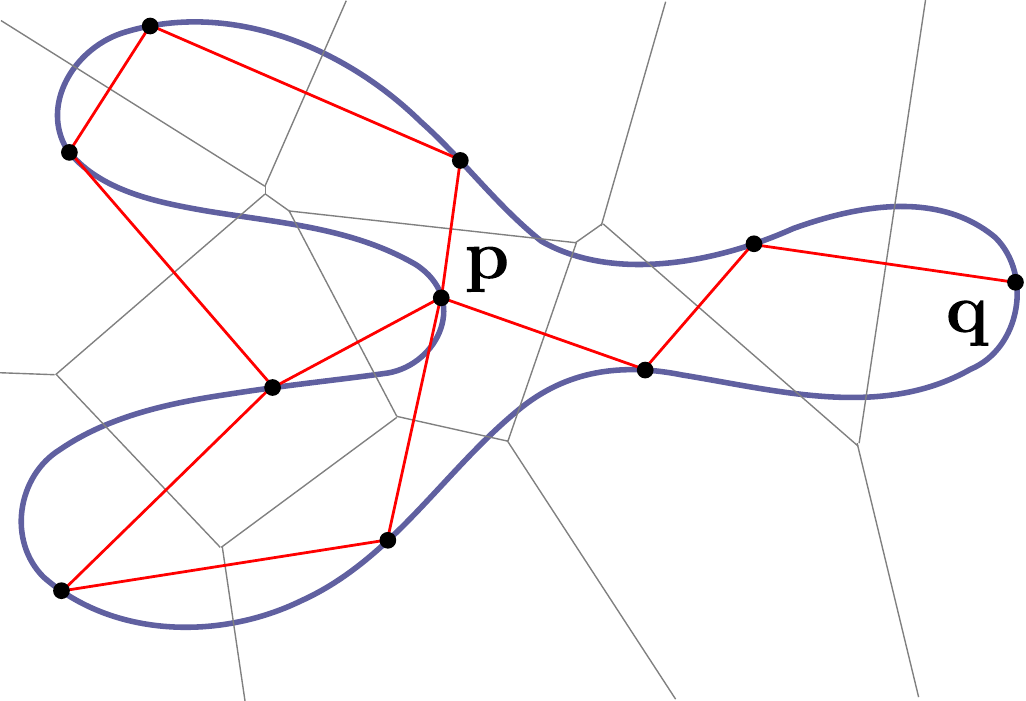}}
  \hspace{.01\linewidth} 
  \subfloat[]{\includegraphics[width=.47\linewidth]{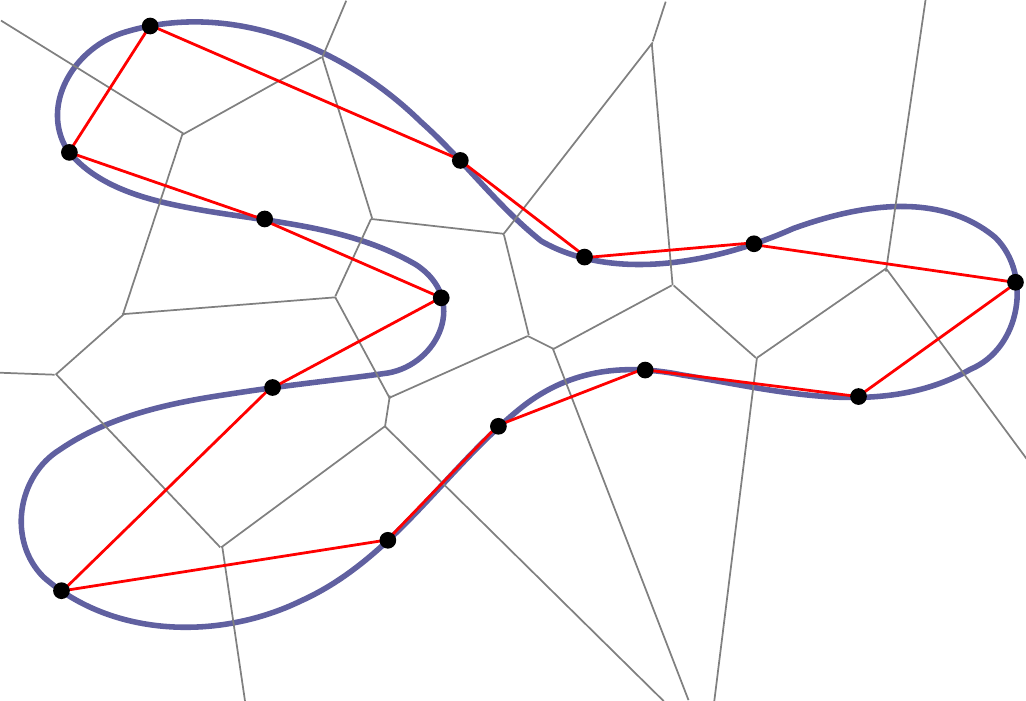}}
  \caption{\label{fig:resDel}(a) the Voronoi complex of a point sample from a
  closed curve and (b) the dual Delaunay graph; (c) the Delaunay graph
  restricted to the curve. In order to make the graph homeomorphic to the
  curve, the sample density has to be increased (d).}
\end{figure}

The \emph{Voronoi cell} of a point $\mathbf{p}$ in a set of points $L$ is:
$$V_{\mathbf{p}} = \{\mathbf{x} \in \mathbb{R}^d \:|\; \|\mathbf{x} -
\mathbf{p}\| \leq \|\mathbf{x} - \mathbf{q}\|, \forall \mathbf{q} \in L,
\mathbf{q} \neq \mathbf{p}\}$$
The intersection of two Voronoi cells in $\mathbb{R}^d$ may be either empty or
a linear face of dimension $d - 1$. Likewise, the intersection of $n$ such
cells is either empty or a linear face of dimension $d - n + 1$, with a minimum
of 0. For example, the intersection of $d$ cells is either empty or an edge
(i.e. dimension 1) and the intersection of $d + 1$ cells is either empty or a
single point (i.e. dimension 0). The Voronoi cell together with the faces of
all dimensions form the \emph{Voronoi complex} of $L$. Fig.
\ref{fig:resDel}(a) shows an example of a Voronoi complex for a set of points
in $\mathbb{R}^2$.

A finite set of point $L$ in $\mathbb{R}^d$ is said to be \emph{non
degenerate} if no $d + 2$ Voronoi cells have a non-empty intersection.

The \emph{Delaunay graph} of a finite set of points $L$ is dual to the Voronoi
complex, in the sense that two points $\mathbf{p}$ and $\mathbf{q}$ in $L$ are
connected in the Delaunay graph if the intersection of the two corresponding
Voronoi cells is not empty. The \emph{Delaunay simplicial complex}
$\mathcal{D}(L)$ is defined by extending the above idea to simplices of higher
dimensions: a face $\sigma$ of dimension $n$ is in $\mathcal{D}(L)$ iff the
intersection of the Voronoi cells corresponding to the $n + 1$ points in
$\sigma$ is non-empty. If the set $L$ is non-degenerate, $\mathcal{D}(L)$ will
contain simplices of dimension at most $d$. Fig. \ref{fig:resDel}(b) shows the
Delaunay graph corresponding to the Voronoi complex in Fig. \ref{fig:resDel}(a).

\subsection{Restricted Delaunay complex}

The concept of \emph{restricted Delaunay graph} was already defined in
\cite{Martinetz-Schulten94}, albeit with a slightly different terminology.

Let $M \subseteq \mathbb{R}^d$ be a manifold of dimension $k$. The
\emph{restricted Voronoi cell} of $\mathbf{p}$ w.r.t a manifold $M$ is:
$$V_{\mathbf{p},\,M} = V_{\mathbf{p}} \cap M$$

The \emph{restricted Delaunay graph} of $L$ with respect to $M$ is a graph
where points $\mathbf{p}$ and $\mathbf{q}$ are connected iff
$V_{\mathbf{p},\,M} \,\cap \,V_{\mathbf{q},\,M} \neq \emptyset$.  Fig.
\ref{fig:resDel}(c) shows the restricted Delaunay graph with respect to a
closed curve. Note that the restricted Delaunay graph is a subset of the
Delaunay graph (Fig. \ref{fig:resDel}(a)) for the same set of points.

The \emph{restricted Delaunay simplicial complex} $\mathcal{D}_{M}(L)$ is the
simplicial complex obtained from the complex of the restricted Voronoi cells
$\mathcal{V}_{M}(L)$.

\subsection{Homeomorphism and $\varepsilon$-sample}

Note that the restricted Delaunay graph in Fig. \ref{fig:resDel}(c) is not
homeomorphic to the closed curve $M$. For instance, point $\mathbf{p}$ has only
one neighbor, instead of two, whereas $\mathbf{q}$ has four. This means that
the piecewise-linear curve $\mathcal{D}_{M}(L)$ has one boundary point and (at
least) one self-intersection and therefore it does not represent $M$ correctly.
 
\begin{figure}[h]
  \centering
  \includegraphics[width=.6\linewidth]{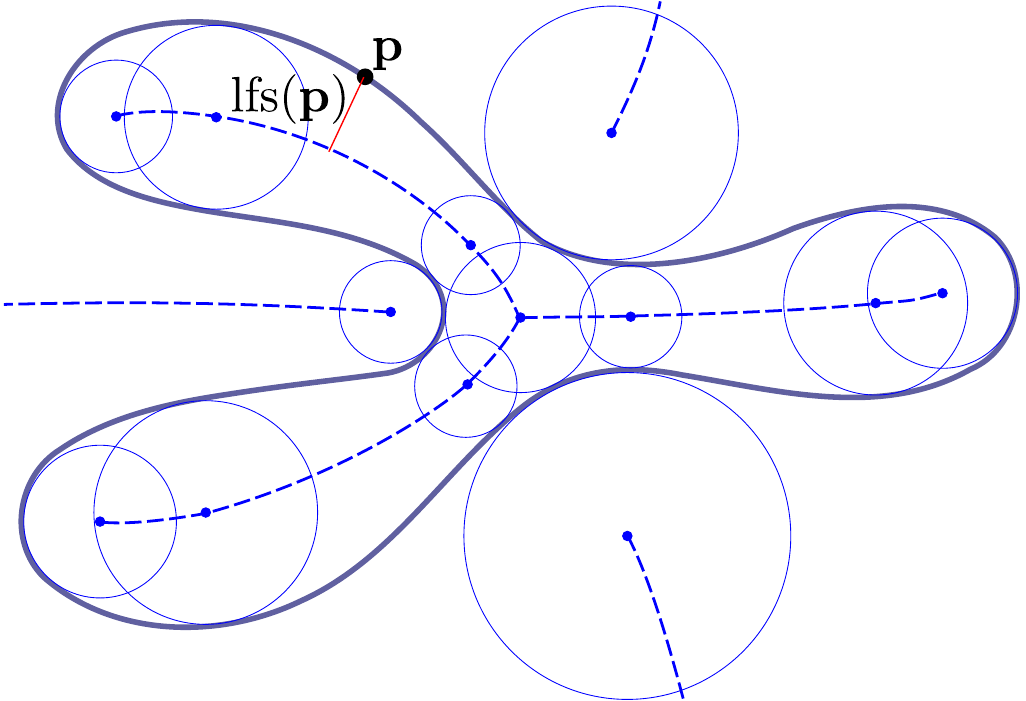}
  \caption{\label{fig:medAx}The medial axis (dashed line) of a closed curve and
  the \emph{local feature size} at a point $\mathbf{p}$.}
\end{figure}
The \emph{medial axis} of $M$ is the closure of the set of points that are the
centers of maximal balls, called \emph{medial balls}, whose interiors are empty
of any points from $M$. The \emph{local feature size} at a point $\mathbf{p}
\in M$ is the distance of $\mathbf{p}$ from the medial axis. The \emph{global
feature size} of a compact manifold $M$ is the infimum of
$\mathrm{lfs}(\mathbf{x})$ over all $M$. Fig. \ref{fig:medAx} describes the
medial axis (dashed blue line) for the manifold in Fig. \ref{fig:resDel}. The
local feature size of point $\mathbf{p}$ in figure is equal to the length of
the red segment.

By definition, the set $L$ is an \emph{$\varepsilon$-sample} of $M$ if for all
points $\mathbf{x} \in M$ there is at least one point $\mathbf{p} \in L$ such that
$\|\mathbf{x} - \mathbf{p}\| \leq \varepsilon \cdot \mathrm{lfs}(\mathbf{x})$.

\begin{thm} \label{thm:eps-cbp}
Under the additional condition that $M$ is \emph{smooth}, there exists an
$\varepsilon > 0$ such that if the set $L$ is an \emph{$\varepsilon$-sample} of $M$, 
then $\mathcal{D}_{M}(L) \approx M$ \cite{Amenta-Bern98}.
\end{thm}

Slightly different values of $\varepsilon$ ensuring the validity of theorem
\ref{thm:eps-cbp} are reported in the literature. It will suffice to our
purposes that such positive values exist. The effect of the above theorem is
exemplified in Fig. \ref{fig:resDel}(d): as the density of $L$ increases the
restricted Delaunay graph becomes homeomorphic to the curve.

\begin{figure}[tcb]
  \centering
  \subfloat[]{\includegraphics[width=.47\linewidth]{restrictedDelaunayHomeomorphic}}
  \hspace{.01\linewidth}
  \subfloat[]{\includegraphics[width=.47\linewidth]{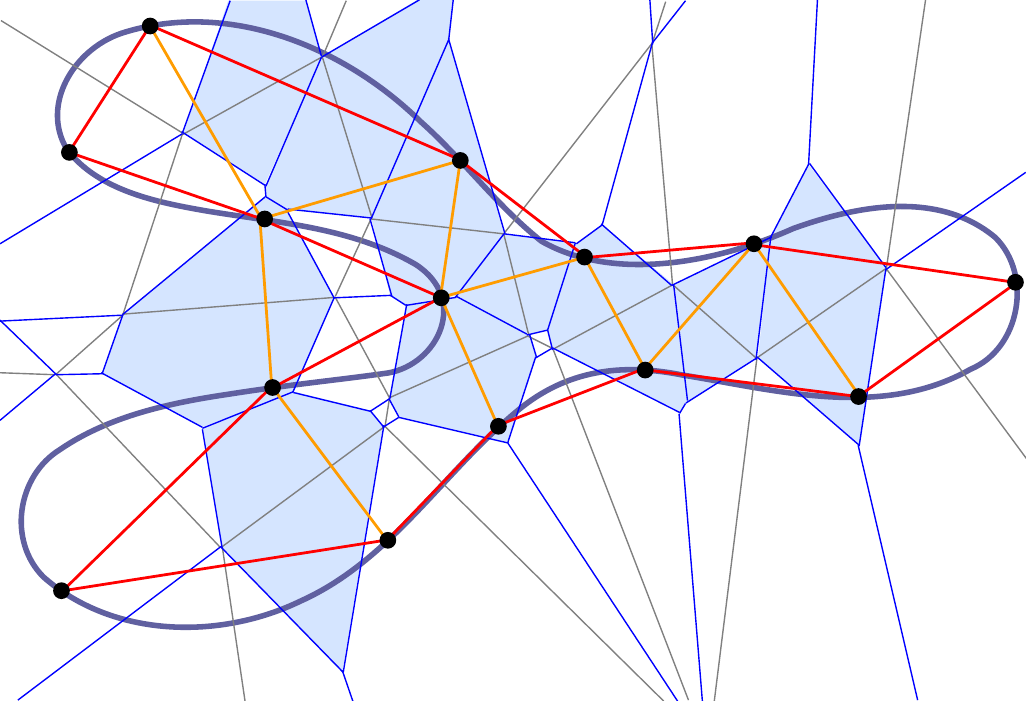}}
  \\
  \subfloat[]{\includegraphics[width=.47\linewidth]{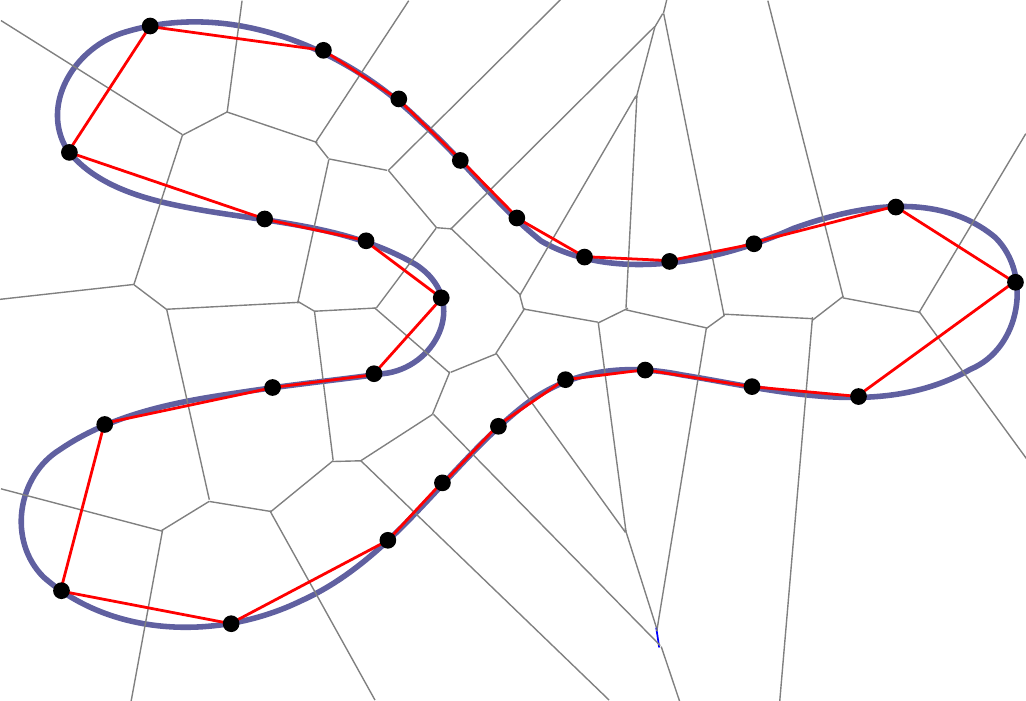}}
  \hspace{.01\linewidth} 
  \subfloat[]{\includegraphics[width=.47\linewidth]{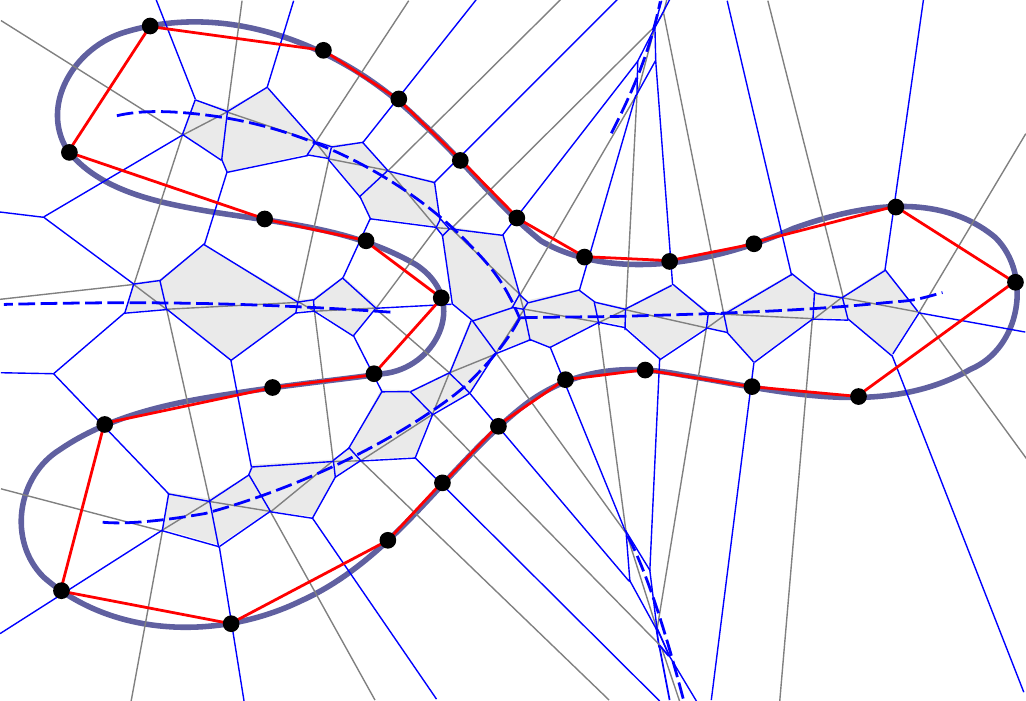}}
  \caption{\label{fig:witness}(a) the restricted Delaunay complex is
  homeomorphic to the curve; (b) the witness complex, however, includes further
  edges, due to the extent of second order Voronoi regions; (c) after
  increasing the sample density, the restricted Delaunay complex remains
  homeomorphic to the curve and (d) coincides with the witness complex.}
\end{figure}
\subsection{Witness complex}

The notion of a \emph{witness complex} is tightly related to the
\emph{competitive Hebbian rule} described in \cite{Martinetz-Schulten94} and,
at least historically, descends from the latter.

Given a set of point $L \in \mathbb{R}^d$, a \emph{weak witness} of a simplex
$\sigma \subset L$ is a point $\mathbf{x} \in \mathbb{R}^d$ for which
$\|\mathbf{x} - \mathbf{a}\| \leq \|\mathbf{x} - \mathbf{b}\|$ for all
$\mathbf{a} \in \sigma$ and $\mathbf{b} \in L - \sigma$. A \emph{strong
witness} is a weak witness $\mathbf{x}$ for which $\|\mathbf{x} -
\mathbf{a}_{i}\| = \|\mathbf{x} - \mathbf{a}_{j}\|$ for all $\mathbf{a}_{i},
\mathbf{a}_{j} \in \sigma$. Note in passing that all points belonging to the
intersection of two Voronoi cells $V_\mathbf{a}$ and $V_\mathbf{b}$ are strong
witnesses for the edge $(\mathbf{a},\mathbf{b})$. The set of all weak witnesses
for this same edge is also called the \emph{second order Voronoi cell} of the
two points $\mathbf{a}$ and $\mathbf{b}$ \cite{Martinetz-etal93}.

The \emph{witness complex} $\mathcal{C}^W(L)$ is the simplicial complex that
can be constructed from the set of points $L$ with a specific set of witnesses
$W$.

\begin{thm} \label{thm:weak-del}
Let $L \subset \mathbb{R}^d$ be a set of points. If every face of a simplex
$\sigma \subseteq L$ has a weak witness in $\mathbb{R}^d$, then $\sigma$ has a
strong witness $\mathbb{R}^d$ \cite{deSilva-Carlsson04} .
\end{thm}

As a corollary, this theorem implies that $\mathcal{C}^W(L) \subseteq
\mathcal{D}(L)$ and in the limit the two complexes coincide, when $W \equiv 
\mathbb{R}^d$.

Theorem \ref{thm:weak-del} does not extend to \emph{restricted} Delaunay
complexes, namely when the witnesses belong to a manifold $M$. A
counterexample is shown in Fig. \ref{fig:witness}(a), which describes
the witness complex constructed from the same $L$ in Fig. \ref{fig:resDel}(d)
using the whole $M$ as the set of witnesses. The second-order Voronoi regions
(in blue) for the connections (in orange) which do not belong to the restricted
Delaunay graph, do intersect the closed curve. This means that $M$ contains
weak witnesses for connections for which $M$ itself contains no strong
witnesses, as the definition of $\mathcal{D}_{M}(L)$ would require.

\begin{thm} \label{thm:weak-res-del}
If $M$ is a compact smooth manifold without boundary of dimension 1 or 2, there
exists an $\varepsilon > 0$ such that if $L$ is an $\varepsilon$-sample of $M$, the
following implication holds: if every face of a simplex $\sigma \subseteq W$
has a weak witness, then $\sigma$ has a strong witness \cite{Attali-etal07}.
\end{thm}

Like before, this implies that $\mathcal{C}^W_M(L)$, i.e. the witness complex
restricted to $M$, is included in $\mathcal{D}_M(L)$ and the two coincide in the
limit, when $W \equiv M$. As a corollary, Theorem \ref{thm:weak-res-del} also
implies that $\mathcal{D}_{M}(L) \approx M$.  Specific, viable values for
$\varepsilon$ are reported in \cite{Attali-etal07}, but once again it will
suffice to our purposes that such positive values exist.

The effect of theorem \ref{thm:weak-res-del} is described in Fig.
\ref{fig:witness}(c) and (d): by further increasing the density of $L$, the
witness complex $\mathcal{C}^W_M(L)$ coincides with $\mathcal{D}_{M}(L)$, when
$W$ is the entire curve. Fig. \ref{fig:witness}(c) shows that, as the density
of $L$ increases, the second-order Voronoi regions for the violating
connections `move away' from $M$ and tend to aggregate around the medial axis. 

The other side of the medal is represented by:

\begin{thm} \label{thm:no-higher-d}
For manifolds of dimension greater than 2, no positive value of
$\varepsilon$ can guarantee that every face of a simplex $\sigma \in
\mathcal{C}_{M}^W(L)$ having a weak witness also has a strong witness
\cite{Oudot2008}.
\end{thm}

\subsection{Finite samples and noise}

As fundamental as it is, Theorem \ref{thm:weak-res-del} is not per se
sufficient for our purposes. First, it assumes the potential coincidence
of $W$ with $M$, which would require an infinite sample, and, second, its proof
requires that $L \subset M$, which is not necessarily true with vector
quantization algorithms and/or in the presence of noise.

An $\varepsilon$-sparse sample $L$ is such that the pairwise distance between
any two points of $L$ is greater than $\varepsilon$. A $\delta$-noisy sample
$L$ is such that no points in $L$ are farther than $\delta$ from $M$.

\begin{thm} \label{thm:finiteSample}
If $M$ is a compact smooth manifold without boundary of dimension 1 or 2,
there exist positive values of $\varepsilon$ and $\delta$ such that if $W$
is a $\delta$-noisy $\delta$-sample of $M$ and $L$ is a $\delta$-noisy,
$\varepsilon$-sparse $(\varepsilon+\delta)$-sample of $M$ then
$\mathcal{C}_{M}^W(L) \subseteq \mathcal{D}_{M}(L)$ \cite{Guibas-Oudot07}.
\end{thm}

\begin{figure}[h]
  \centering
  \includegraphics[width=.28\linewidth]{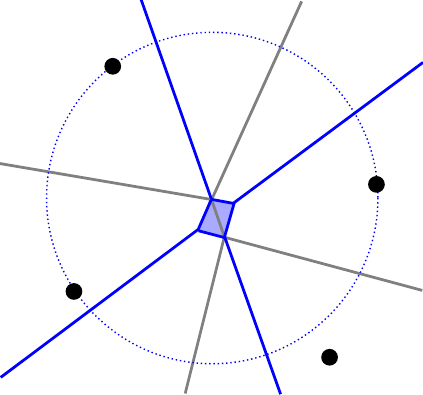}
  \hspace{.02\linewidth}
  \includegraphics[width=.28\linewidth]{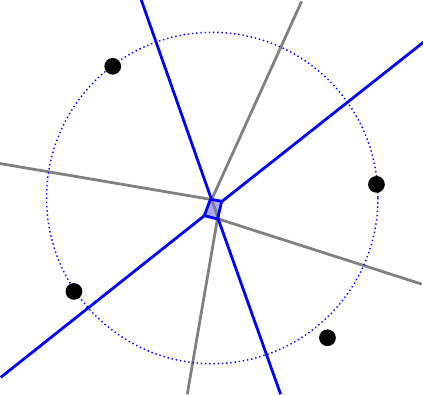}
  \hspace{.02\linewidth}
  \includegraphics[width=.28\linewidth]{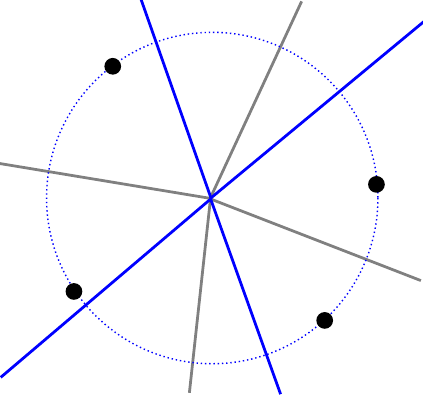}
  \hspace{.02\linewidth}
  \caption{\label{fig:cocircular} As the points approach co-circularity, the
  second order Voronoi region in the midst becomes arbitrarily small.}
\end{figure}

Actually, a stronger result holds in the case of curves, stating that
$\mathcal{C}_{M}^W(L)$ and $\mathcal{D}_{M}(L)$ will coincide. For surfaces
this guarantee cannot be enforced due to a problem that is described in Fig.
\ref{fig:cocircular}. Even if $L$ is non-degenerate, in fact, it may contain
points that are arbitrarily close to a co-circular configuration, thus making the area
of the second-order Voronoi region in the midst so small that no density
condition on $W$ could ensure the presence of an actual witness. This means
that, in a general and non-degenerate case, a few connections
will be missing from $\mathcal{C}^W(L)$.

\section{Self-Organizing Adaptive Map (SOAM)}

The intuitive idea behind the SOAM algorithm introduced here, which is derived
from the GWR algorithm \cite{Marsland-etal02}, is pretty simple. Points like
$\mathbf{p}$ and $\mathbf{q}$ in Fig.\ref{fig:resDel}(c), whose neighborhoods
are not of the expected kind, are true symptoms of the fact that network
structure, i.e. the witness complex $\mathcal{C}_{M}^W(L)$, is not
homeomorphic to the input manifold $M$. Symptoms like these are easily
detectable, provided that the expected dimension of $M$ is known. Then, in the
light of the above theoretical results, the network might react by just
increasing the density of its units. Given that the target density is function
of the \emph{local} feature size of $M$, new units need only be added where
and when required.

\begin{figure}[ht] 
  \centering
\[\begin{xy} /r10mm/:
  ,0="p1" 			,(-.6,0)*@{o}="v0";
  					 (0,0)*@{*}="v1"**@{-};
  					 (.6,0)*@{o}="v2"**@{-};
  ,+/r30mm/="p2"	,"v0"+"p2"*@{*};
  					 "v1"+"p2"*@{*}**@{-};
  					 "v2"+"p2"*@{*}**@{-};
  ,+/r30mm/="p3"	,"v0"+"p3"*@{*};
  					 "v1"+"p3"*@{o};
  					 "v2"+"p3"*@{*};
  ,"p1"-(0,1.3) 	,{*@{*}\xypolygon6{~:{(.8,0):}~<{-}~>{.}@{o}}}
  ,"p2"-(0,1.3)		,{*@{*}\xypolygon6{~:{(.8,0):}~<{-}@{*}}}
  ,"p3"-(0,1.3)		,{*@{o}\xypolygon6{~:{(.8,0):}@{*}}}
  ,"p1"-(0,2.2)+D*+!U\txt{$\mathrm{St}(u_i)$}
  ,"p2"-(0,2.2)+D*+!U\txt{$\mathrm{Cl}(\mathrm{St}(u_i))$} 
  ,"p3"-(0,2.2)+D*+!U\txt{$\mathrm{Lk}(u_i)$} 
  \end{xy}\]
  \caption{\label{fig:stcllnk}The \emph{star}, \emph{closure} and \emph{link}
  of a vertex in a simplicial complex of dimensions 1 (above) and 2 (below).}
\end{figure}
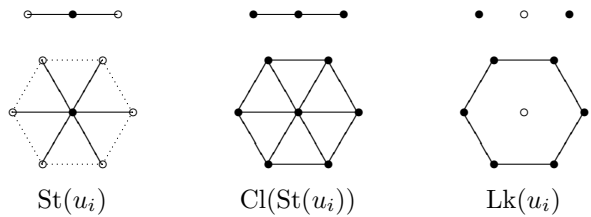
\subsection{Topology-driven state transitions}

The \emph{state} of units in a SOAM is determined according to the topology of
their neighborhoods. Given a unit $u_i$, its neighborhood in a simplicial
complex is defined as the \emph{star} $\mathrm{St}(u_i)$, consisting of $u_i$
together with the edges and triangles that share $u_i$ as a vertex. The
\emph{closure} of a star $\mathrm{Cl}(\mathrm{St}(u_i))$ is obtained by adding
all simplices in $\mathcal{C}(L)$ having a non-null intersection with
$\mathrm{St}(u_i)$. Finally, the \emph{link} is then defined as
$$\mathrm{Lk}(u_i)=\mathrm{Cl}(\mathrm{St}(u_i)) - \mathrm{St}(u_i)$$
The eight possible states for units in a SOAM are:
\begin{itemize}[\settowidth\labelwidth{\emph{habituated}}]
  \item[\emph{active}]The default state of any newly-created unit.
  \item[\emph{habituated}]The value of the firing counter $f$ (see below) of the
  unit is greater than a predefined threshold $T_f$.
  \item[\emph{connected}]The unit is \emph{habituated} and all the units in
  its link are \emph{habituated} as well.
  \item[\emph{half-disk}]The \emph{link} of unit is homeomorphic to an
  half-sphere.
  \item[\emph{disk}]The \emph{link} of unit is homeomorphic to a sphere.
  \item[\emph{boundary}]The unit is an \emph{half-disk} and all its neighbors
  are \emph{regular} (see below).
  \item[\emph{patch}]The unit is a \emph{disk} and all its neighbors are
  \emph{regular}.
  \item[\emph{singular}]The unit is over-connected, i.e. its link exceeds a
  topological sphere. More precisely, the link contains a sphere \emph{plus}
  some other units and, possibly, connections.
\end{itemize}
In the case of dimension 2, a unit whose link is a disk but contains three units
only is a \emph{tetrahedron} and, by definition, is considered \emph{singular} as well.

Note that all the above topological conditions can be determined
combinatorially and hence very quickly. On the other hand, the actual test to
be performed depends on the expected dimension of the input manifold $M$. In
Fig. \ref{fig:stcllnk} on the right, the two links above and below are
\emph{disk}s in dimension 1 and 2, respectively. For instance, in a
complex of dimension 2, a link like the one above on the right in Fig.
\ref{fig:stcllnk} would not be a \emph{disk} at all.

Two further, derived definitions will be convenient while describing the
SOAM algorithm:
\begin{itemize}[\settowidth\labelwidth{\emph{habituated}}]
  \item[\emph{regular}]The unit is in one of these states: \emph{half-disk},
  \emph{disk}, \emph{boundary} or \emph{patch}.
  \item[\emph{stable}]In this version of the algorithm, only the \emph{patch}
  state is deemed \emph{stable}.
\end{itemize}
\begin{figure}[ht]
  \centering
  \includegraphics[width=.8\linewidth]{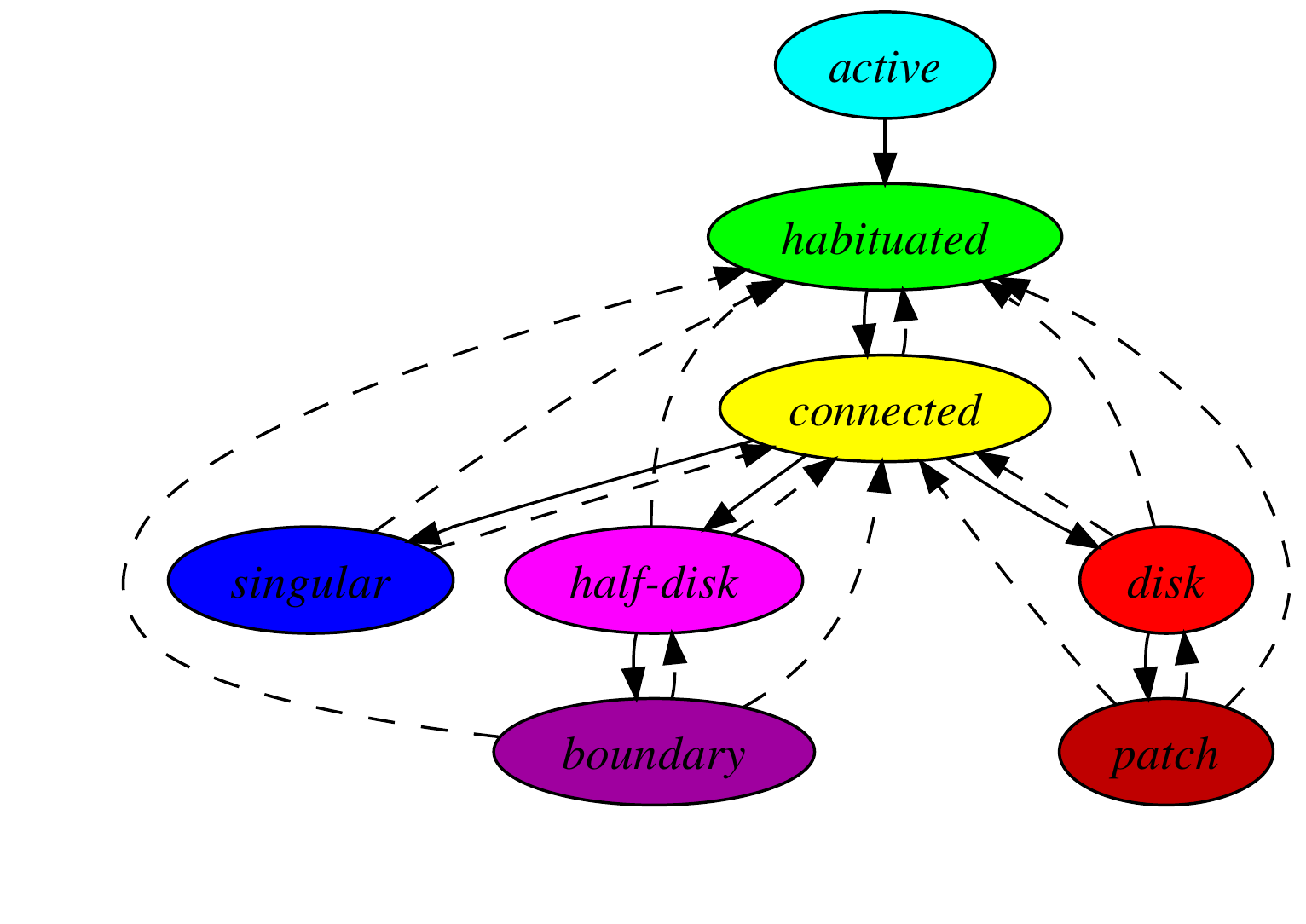}
  \caption{\label{fig:st-trans}Topology-driven state transitions for units in a
  SOAM. The same state colors are used in other figures.}
\end{figure}

\subsection{Adaptive insertion thresholds}

\begin{figure}[hb]
  \centering
  \includegraphics[width=1\linewidth]{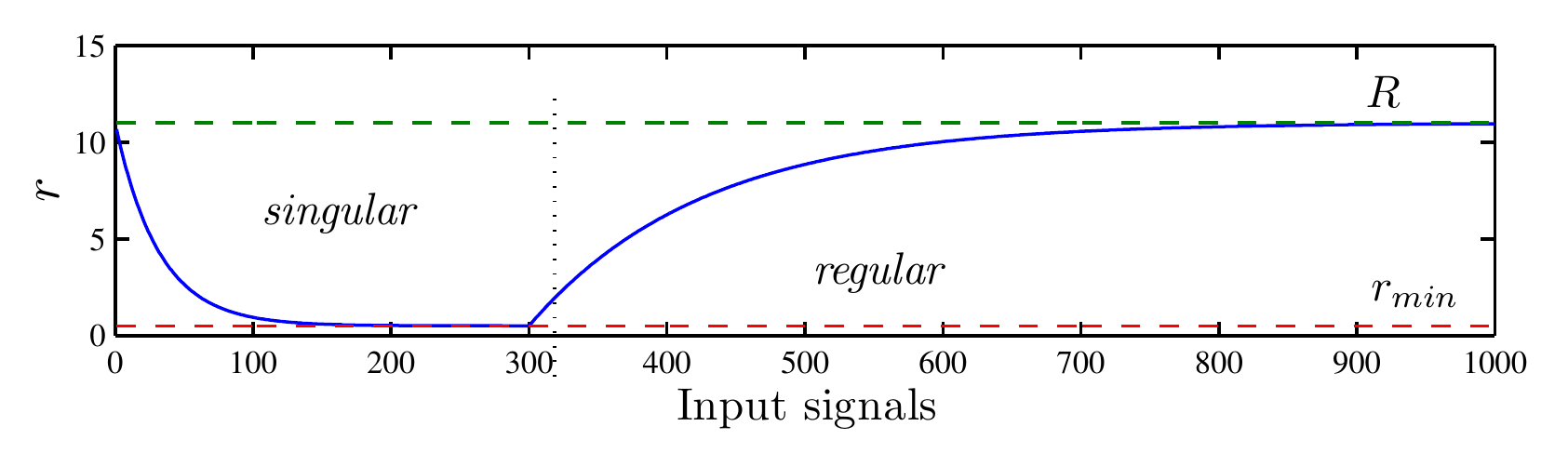}
  \caption{\label{fig:hab-dis}The overall model of habituation and
  dishabituation of insertion thresholds, with different time constants $\tau$.}
\end{figure}
The model adopted in the GWR algorithm for the exponential decay of the firing
counter $f$ has been derived from the biophysics of the \emph{habituation} of
neural synapses \cite{Marsland-etal02}. The equation that rules the model is:
\begin{equation}\label{eqn:hab}
h(t) = H - \frac{1}{\alpha}(1 - \mbox{e}^{-\alpha \cdot t / \tau})
\end{equation}
where $h(t)$ is the value being adapted, as a function of time $t$, $H$ is
the maximum, initial value and $\alpha$ and $\tau$ are suitable parameters.
Equation (\ref{eqn:hab}) is the solution of the differential equation:
\begin{equation}\label{eqn:d-hab}
\frac{\mbox{d}h(t)}{\mbox{d}t} =\frac{\alpha \cdot (H - h(t)) - 1}{\tau}
\end{equation}

The reverse model, of \emph{dishabituation}, is also considered here:
\begin{equation}\label{eqn:dis}
h(t) = H - \frac{1}{\alpha}\cdot\mbox{e}^{(-\alpha \cdot t) / \tau)}
\end{equation}
which in turn is the solution of:
\begin{equation}\label{eqn:d-hab}
\frac{\mbox{d}h(t)}{\mbox{d}t} =\frac{\alpha}{\tau} \cdot (H - h(t))
\end{equation}

In the SOAM algorithm, the overall model of habituation and dishabituation is
adopted for adapting the local insertion thresholds $r$ of units (see below).
More precisely, the value of $r$ for \emph{singular} units decays exponentially
to a value $r_{min}$, while the state persists, whereas the value of $r$ of
\emph{regular} units grows asymptotically to the initial value $R$.

\begin{figure}[hctb]
  \centering
  \includegraphics[width=.48\linewidth]{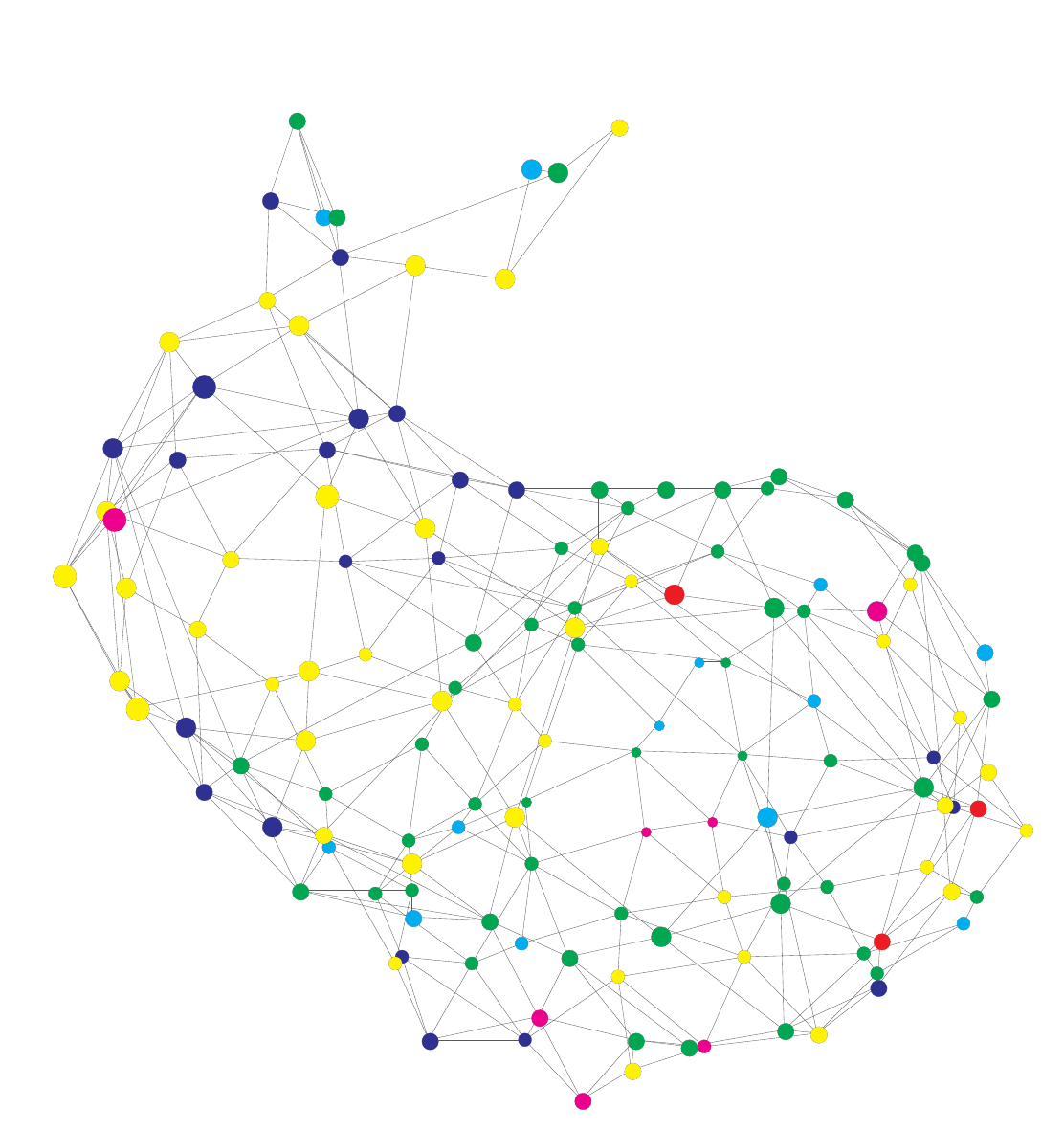}
  \includegraphics[width=.48\linewidth]{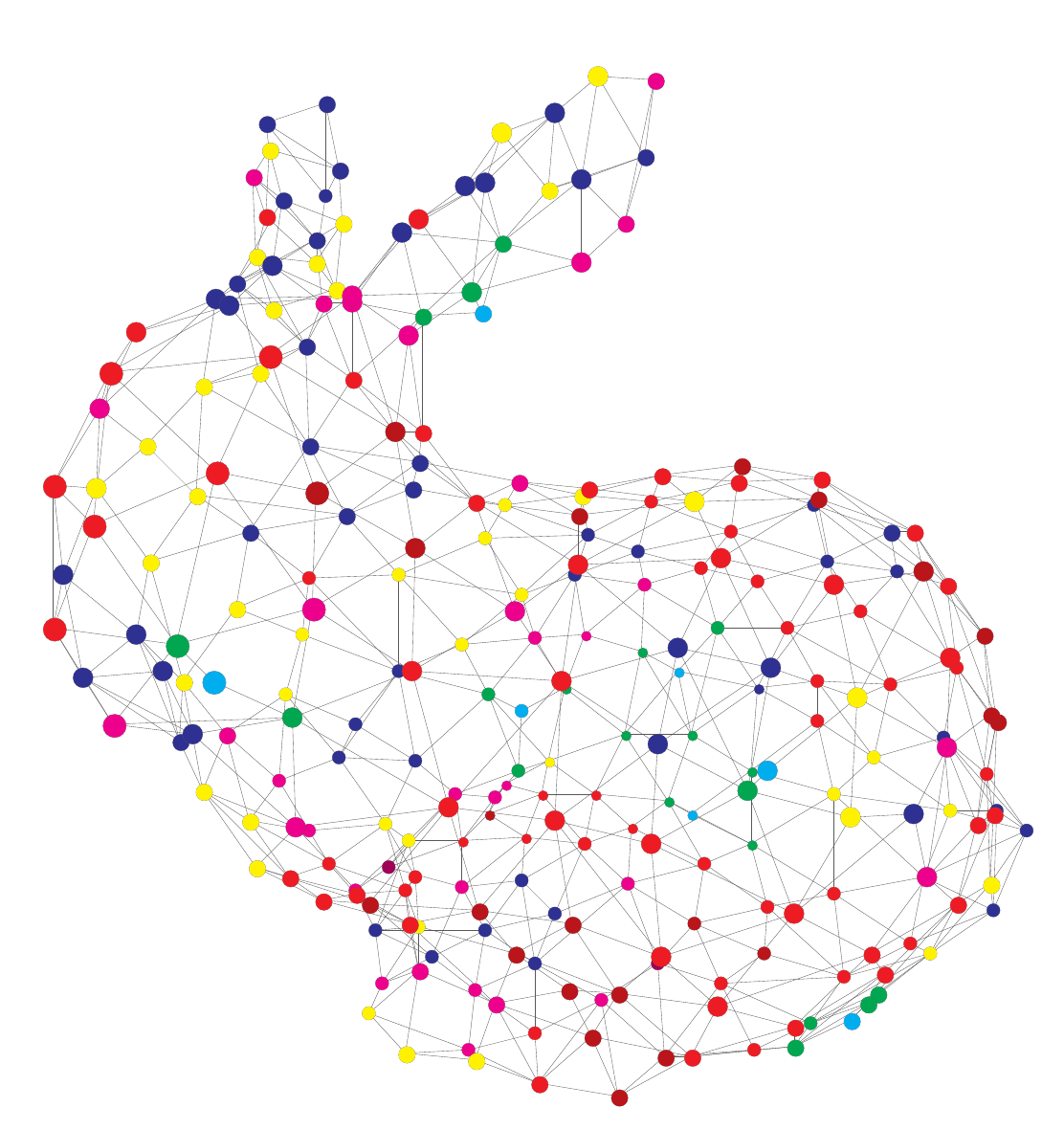}
  \includegraphics[width=.48\linewidth]{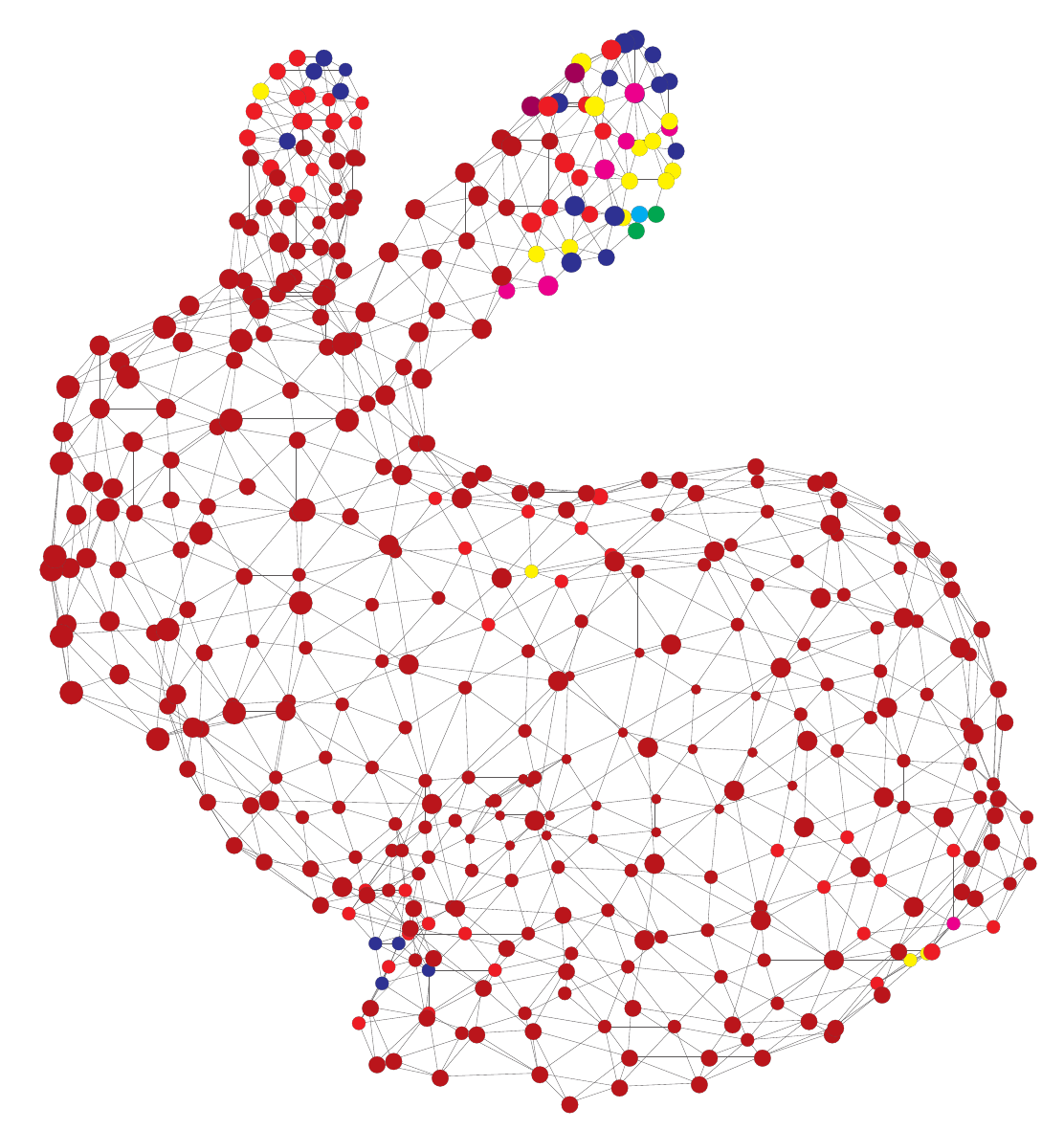}
  \includegraphics[width=.48\linewidth]{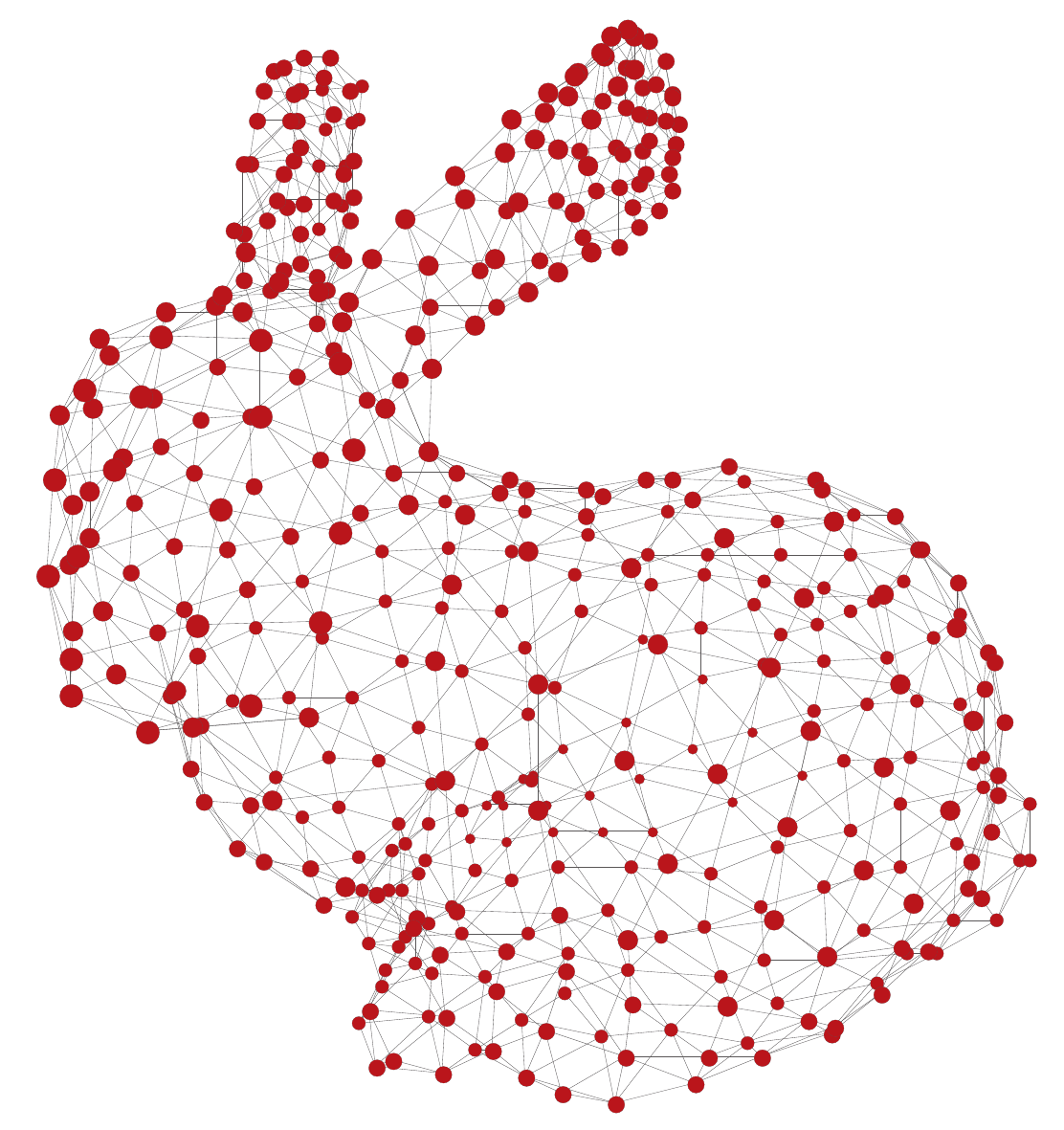}
  \caption{\label{fig:seq-bunny}
  	A SOAM reconstructing the the Stanford bunny. In the final structure, all
  	435 units are in the \emph{stable} state.}
\end{figure}
\subsection{The SOAM algorithm}

Initially, the set $L$ of units contains two units $u_0$ and $u_1$ only, with
positions assigned at random according to $P(\mathbf{\xi})$. The $C$ of connections
is empty. All firing counters $f$ and insertion thresholds $r$ of newly
created units are initialized to $F$ and $R$, respectively.
\begin{enumerate}
  \item Receive the next sample $\mathbf{\xi}$ from the input stream.
  \item Determine the indexes $b$ and $s$ of the two units that are closest and
  second-closest to $\mathbf{\xi}$, respectively
  \begin{eqnarray*}
    \mathbf{p}_b &=& \arg \min_{u_{i} \in A} \|\mathbf{\xi} - \mathbf{p}_{i}\|
    \\ \mathbf{p}_s &=& \arg \min_{u_{i} \in A - {u_{b}}} \|\mathbf{\xi} -
    \mathbf{p}_{i}\|
  \end{eqnarray*}
  \item Add connection $(b,s)$ to $C$, if not already present, otherwise set its
  $age$ to $0$.
  \item Unless $u_b$ is \emph{stable}, increase by one the $age$ of all its
  connections. Remove all connections of $u_b$ whose $age$ exceeds a given
  threshold $T_{age}$. Then, remove all units that remain isolated.
  \item If $u_b$ is at least \emph{habituated} and $\|\mathbf{\xi} -
  \mathbf{p}_{b}\| > r_{b}$, create a new unit $u_{n}$:
  \begin{itemize}
    \item Add the new unit $u_{n}$ to $L$
    \item Set the new position: $\mathbf{p}_{n} = (\mathbf{p}_{b} +
    \mathbf{\xi}) / 2$
    \item Add new connections $(b,n)$ $(n,b)$ to $C$
    \item Remove the connection $(b,s)$ from $C$
  \end{itemize}
  Else, if $u_b$ is at least \emph{habituated} and $\|\mathbf{p}_{s} -
  \mathbf{p}_{b}\| < r_{min}$, merge $u_{b}$ and $u_{s}$:  
  \begin{itemize}
    \item Set the new position: $\mathbf{p}_{b} = (\mathbf{p}_{b} +
    \mathbf{p}_{s}) / 2$
    \item For all connections $(s,x)$, add a new connection $(b,x)$ to $C$, if
    not already present
    \item Remove all connections $(s,x)$ from $C$
    \item Remove $u_{s}$ from $L$
  \end{itemize}
  \item Adapt the firing counters of $u_{b}$ and of all units $u_{nb}$ that are 
  connected to $u_{b}$
  \begin{eqnarray*}
  	\Delta f_{b} &=& (\alpha_{h} \cdot (F - f_{b}) - 1) / \tau_{f} \\
  	\Delta f_{nb} &=& (\alpha_{h} \cdot (F - f_{nb}) - 1) / \tau_{f,\,n}
  \end{eqnarray*}
  \item Update the \emph{state} of $u_{b}$, according to the value of
  $f_{b}$ and the topology of its neighborhood of connected units.
  \item If the unit $b$ is \emph{singular}, adapt its insertion threshold
  $$ \Delta r_{b} = (\alpha_{r} \cdot (R - r_{b}) - 1) / \tau_{r,\,hab}$$
  Else, if the unit $b$ is \emph{regular}, adapt its insertion threshold in the
  opposite direction
  $$ \Delta r_{b} = ((\alpha_{r} / \tau_{r,\,dis}) \cdot (R - r_{b})$$
  \item Unless $u_{b}$ is \emph{stable}, adapt its positions and those of all
  connected units $u_{nb}$
  \begin{eqnarray*}
    \Delta\mathbf{p}_{b} &=& \eta_{b} \cdot f_{b} \cdot (\mathbf{\xi} -
    \mathbf{p}_{b}) \\
    \Delta\mathbf{p}_{nb} &=& \eta_{nb} \cdot f_{nb}
    \cdot (\mathbf{\xi} - \mathbf{p}_{nb})
  \end{eqnarray*}
  Else, if $u_{b}$ is \emph{stable}, adapt only the position of $u_{b}$ itself
  $$ \Delta\mathbf{p}_{b} = \eta_{stable} \cdot f_{b} \cdot (\mathbf{\xi} -
    \mathbf{p}_{b}) $$
  where, typically, $\eta_{stable} \ll \eta_{b}$.
  \item If further inputs are available, return to step (1), unless some
  termination criterion has been met.
\end{enumerate}

\begin{figure}[t]
  \centering
  \includegraphics[width=.48\linewidth]{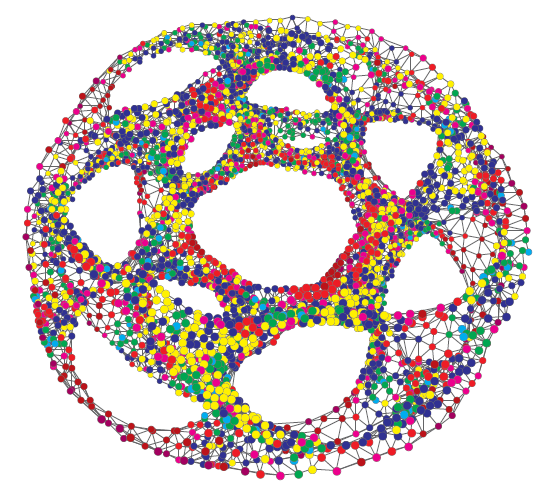}
  \includegraphics[width=.48\linewidth]{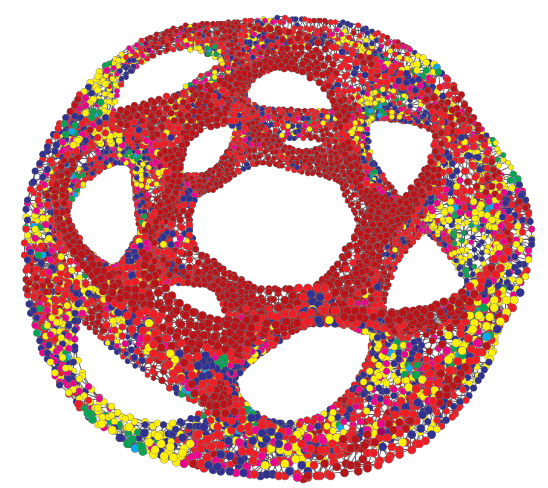}
  \includegraphics[width=.48\linewidth]{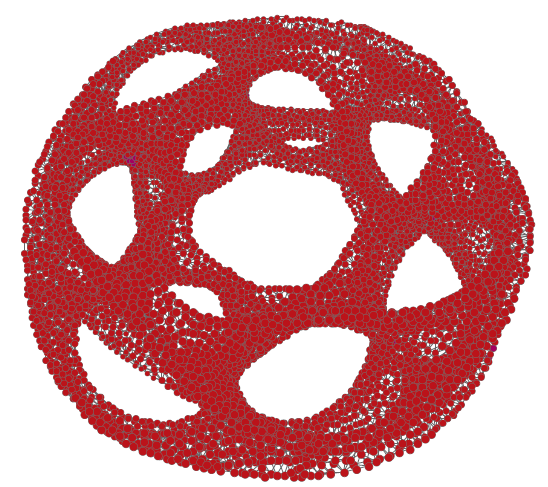}
  \includegraphics[width=.48\linewidth]{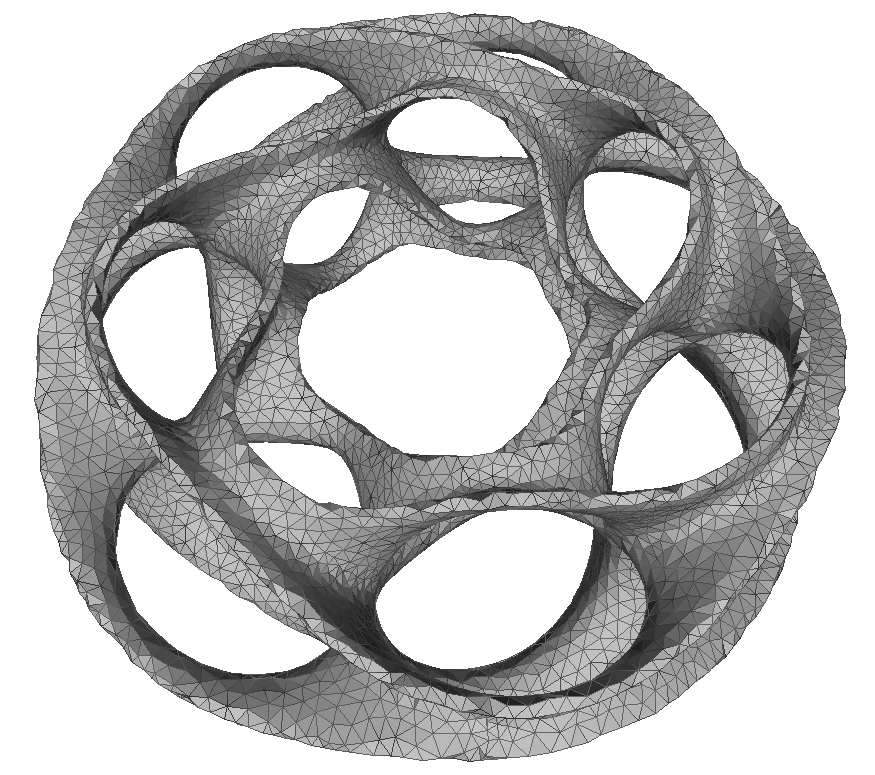}
  \caption{\label{fig:heptoroid}A SOAM reconstructing the heptoroid, which has
  \emph{genus} 22. The final structure includes 10,103 units and has been
  obtained after some 8M input signals.}
\end{figure}
\section{Experimental evidence}

Practical experiments have been performed to assess the properties of the SOAM
algorithm, in particular with surfaces, as this is by far the trickiest case.
The test set adopted includes 32 triangulated meshes, downloaded from the
AIM@SHAPE repository \cite{Falcidieno04}. The reason for this choice is that the
homeomorphism with a triangulated mesh can be easily verified. In fact, two
closed, triangulated surfaces are homeomorphic if they are both orientable (or
non-orientable) and have the same Euler characteristic \cite{Zomorodian05}:
$$\chi = vertex - edges + faces$$
Meshes in the test set have \emph{genus} (i.e. the number of tori whose
connected sum is homeomorphic to the given surface \cite{Zomorodian05}),
ranging from 0 to 65. All meshes were rescaled to make each bounding box 
have a major size of 256. In the experiments, mesh vertices were selected at
random, with uniform distribution.

For completeness, a few test have also been performed with meshes with simple
boundaries and with non-orientable surfaces (e.g. the Klein bottle), directly
sampled from parametric expressions.

\begin{figure}[h]
  \centering
  \includegraphics[width=1\linewidth]{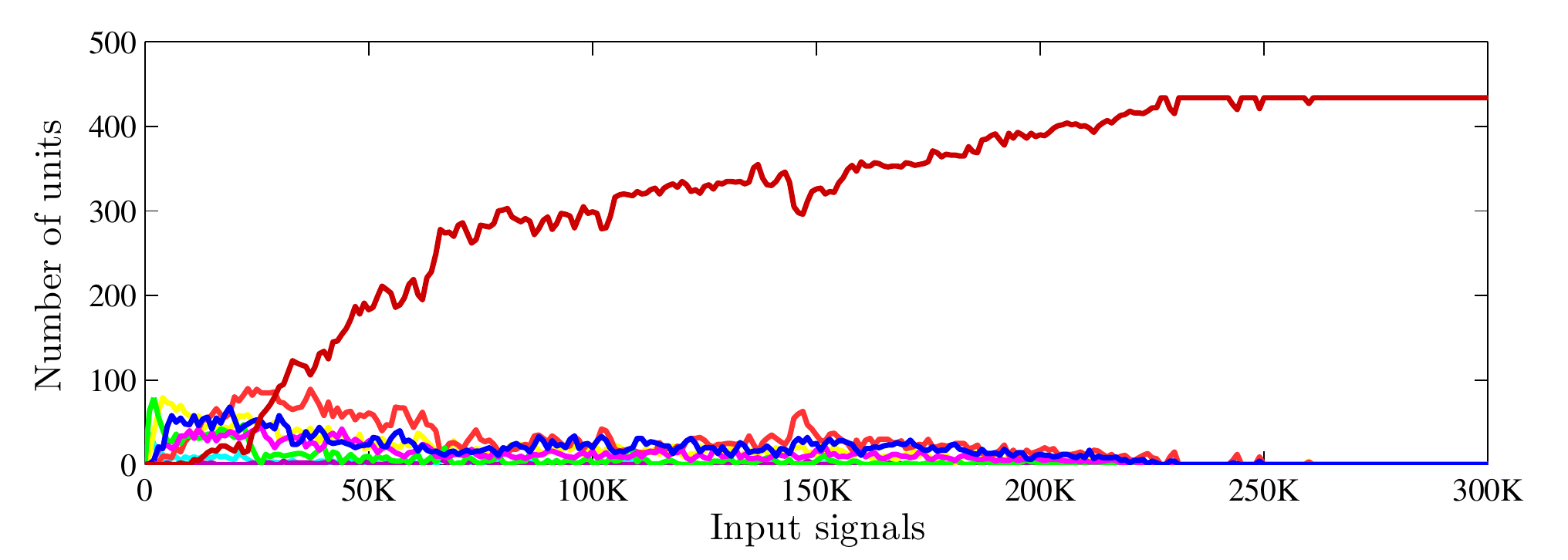}
  \caption{\label{fig:bunny-stat}Typical run of the SOAM algorithm for the
  Stanford bunny reconstruction. Each line charts the number of units in the
  corresponding state/color.}
\end{figure}
\subsection{Topological convergence}

For the whole set of triangulated meshes and in the absence of noise, it has
been possible to make the SOAM reach a condition where all units were
\emph{stable} and the resulting structure was homeomorphic to the input mesh.
Better yet, it has been possible to identify a common set of algorithm
parameters (apart from the dimension of $M$) that, under conditions to be
clarified below, proved to be adequate to the whole test set.

The typical, observed run of the adaptation process is charted in Fig.
\ref{fig:bunny-stat}, which refers to the input mesh of 72,027 vertices
for the process in Fig. \ref{fig:seq-bunny}. As we can see, the number of
\emph{stable} units (dark red line) in the SOAM grows progressively until a
level of equilibrium is reached and maintained from that point on. In contrast,
\emph{singular} nodes (blue line) tend to disappear, although occasional onsets
may continue for a while, for reasons that will be explained in short. Needless
to say, in the most complex test cases a similar level of equilibrium was
reached only after a much larger number of signals, but substantially the same
behaviour has been observed in all successful runs.

\subsection{Algorithm parameters and their effect}

All values for algorithm parameters reported here belong to the common set
that has been identified empirically.

The values $\eta_{b}=0.05$ and $\eta_{nb}=0.0005$ for unit positions are
slightly higher than those suggested in the literature, as it emerged that an
increased mobility of units in pre-\emph{stable} states accelerates the
process. Actually, this accentuated mobility also proved to be the way the
problem described in Fig. \ref{fig:cocircular} could be solved. In fact, only
with much lower values of $\eta_{b}$ and $\eta_{nb}$ the occasional small
`holes' in a diamond of four units became persistent and could not be closed.
A possible explanation, to be fully investigated yet, is that the higher
mobility of units acts like a `simulation of simplicity'
\cite{Edelsbrunner-Mucke88}, that is, a perturbation that makes
low-probability configurations ineffective.

The values $\tau_{f}=3.33$ and $\tau_{f,\,n}=14.33$ for firing counters,
with threshold $T_{f}=0.243$ were chosen initially and never changed afterwards.

In contrast, the values $\tau_{r,\,hab}=3$ for insertion thresholds require
some care, as a shorter time constant may produce an excessive density of units
and a longer one may slow down the process significantly. In addition, the
appropriate combination with the values $\tau_{r,\,dis}=9$, $T_{age}=30$
$\eta_{stable}=0.02$ solves a problem with the connection aging mechanism.
The aging mechanism, in fact, is meant to cope with the mobility of units
during the adaptation process: any connection becoming unsupported by
witnesses will age rapidly and eventually be removed. The drawback is that it
acts in probability: in general, the probability of removing a connection is
inversely proportional to the probability of sampling a supporting witness,
which is not the same as having no witness at all. In all borderline cases,
which are particularly frequent when - once again - four points approach
co-circularity, connections are removed and created in a sort of `blinking'
behavior.

\begin{figure}[ht]
  \centering
  \mbox{} \hfill
  \includegraphics[width=.28\linewidth]{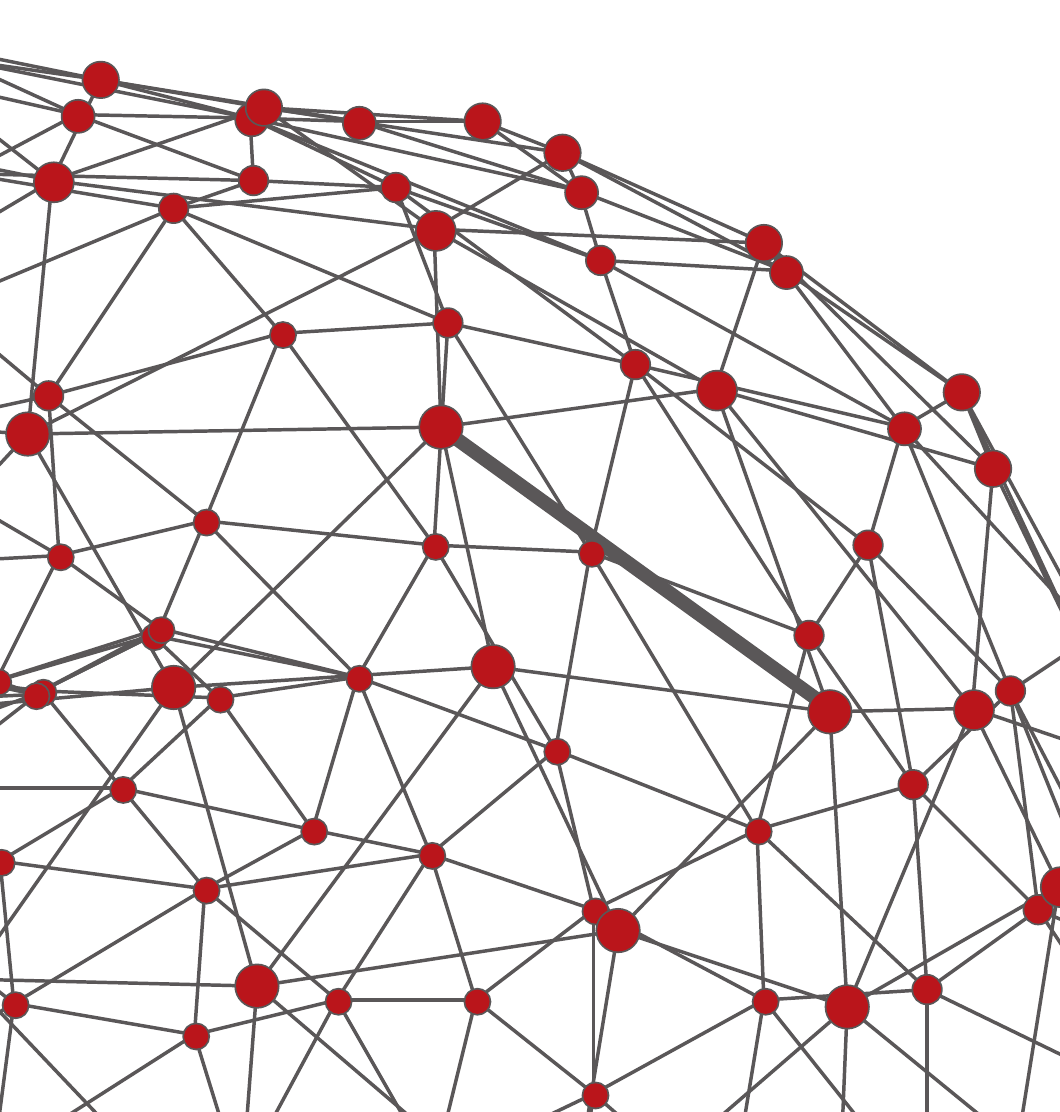}
  \hfill 
  \includegraphics[width=.28\linewidth]{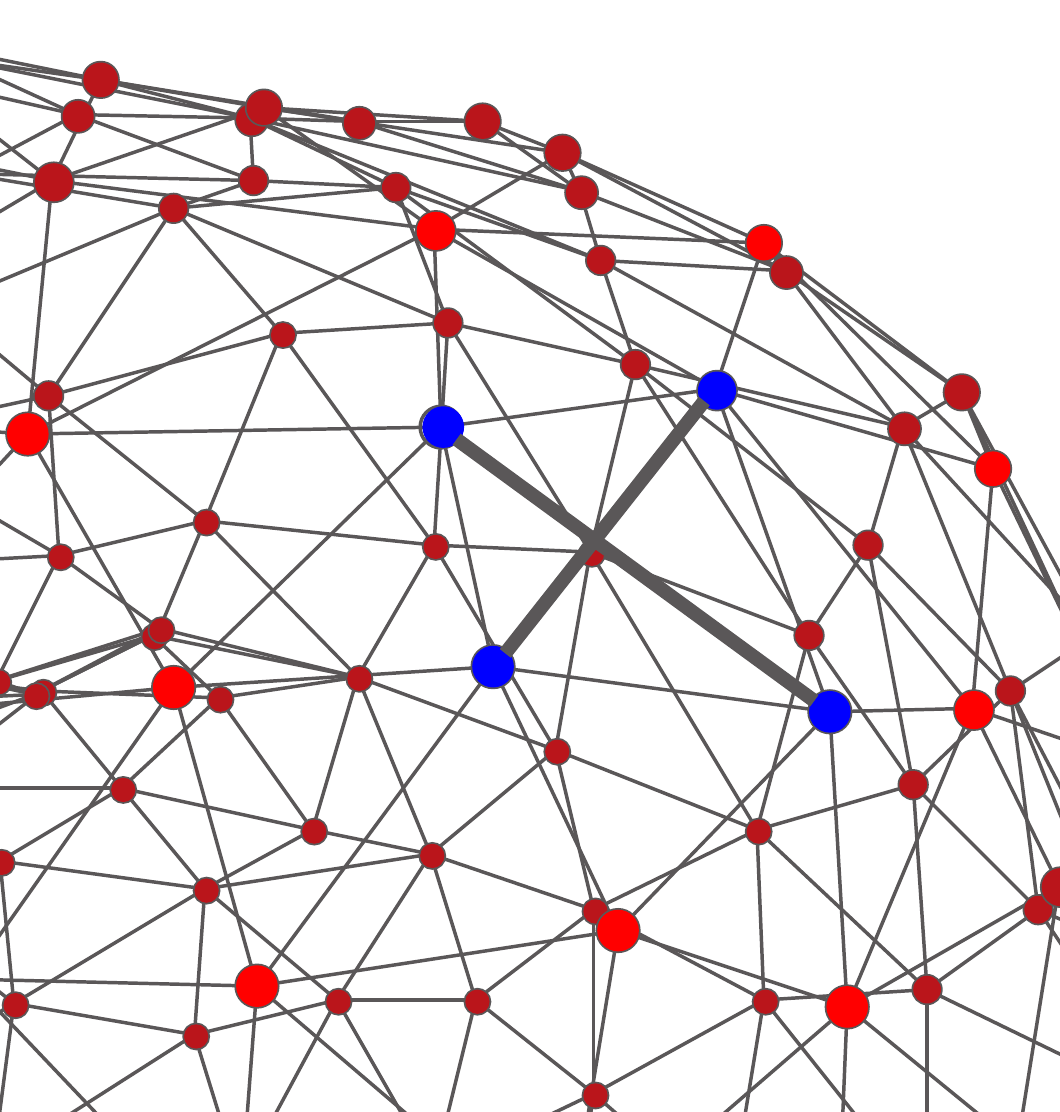}
  \hfill
  \includegraphics[width=.28\linewidth]{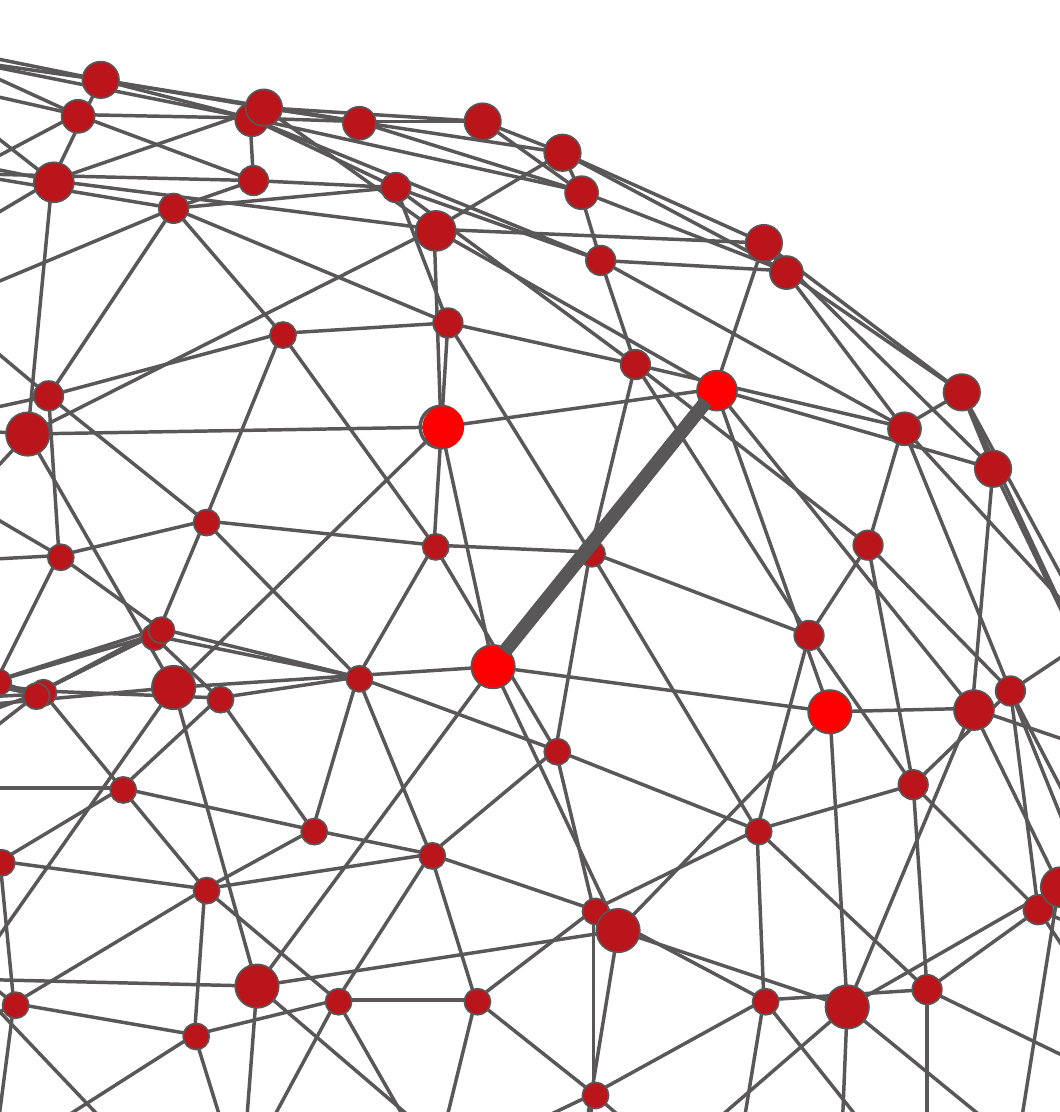}
  \hfill \mbox{} 
  \caption{\label{fig:edg-flp}\emph{Edge flipping} in a SOAM.}
\end{figure}
In step 4 of the algorithm, the aging mechanism is simply stopped for
connections between \emph{stable} units. From that point on, any 
`counter-witness' to the correctness of the triangulation will create
an extra connection that turns a set of four points into a \emph{singular}
tetrahedron, also resuming the aging mechanism. The violating connection
will be removed eventually and the \emph{stable} state will be established
again. Overall, the resulting effect, depicted in Fig. \ref{fig:edg-flp}, is very
similar to \emph{edge flipping} \cite{Edelsbrunner06}. With an appropriate
choice of the above values, occasional edge-flippings will not increase the
density of units. Episodes of this kind may continue to occur after the network
has first reached stability.

As one may expect, the maximum and minium insertion thresholds, $R$ and
$r_{min}$ are critical parameters, since they rule the properties of the the
SOAM as an $\varepsilon$-sample of the input manifold. Surprisingly however,
setting those values proved less critical in practice. With a proper setting
of $\tau_{r,\,hab}$, the density of units in the SOAM increases only when and
as required, so that $r_{min}$ just acts as a lower threshold which is seldom
reached, as proved by the fact that the merge in step 5 in the algorithm
occurred very rarely. This could also be due to the entropic attitude of the
algorithm, which promotes unit sparseness. The adopted value $r_{min}=0.5$ is
equal to twice the maximum distance between neighboring points in the input
data sets.

The parameter $R$ defines the minimum network density and hence the desired
quality of the reconstruction. On the other hand, for reasons not yet
completely clear from a theoretical standpoint, $R$ played no role in
determining convergence. For instance, the value $R=25$, i.e. 10\% of the
major size of any bounding box, proved adequate in all cases. A possible
explanation is that the high mobility of units makes it highly improbable the
occurrence of false positives, intended as stable network configurations that
do not represent $M$. Perhaps due to similar reasons, adapting the insertion
threshold also of \emph{boundary} units proved unnecessary, at least with
closed surfaces, to ensure convergence.

\begin{figure}[htb]
  \centering
  \includegraphics[width=.40\linewidth]{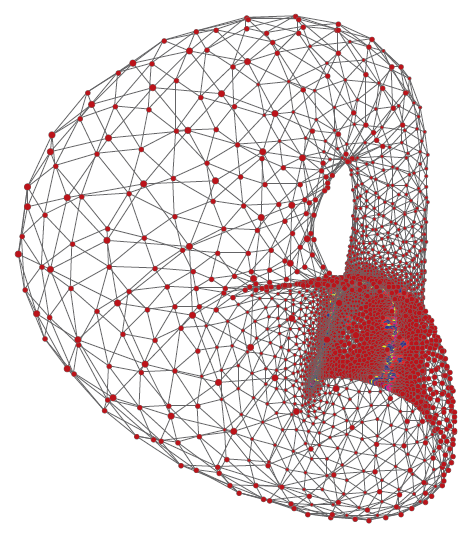}
  \hspace{.1\linewidth}
  \fbox{\includegraphics[width=.35\linewidth]{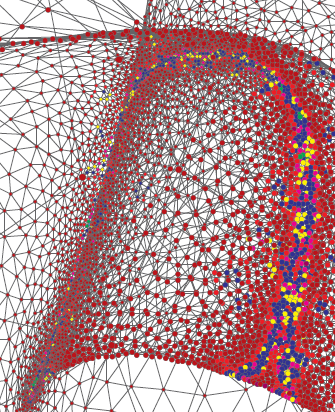}}
  \includegraphics[width=.40\linewidth]{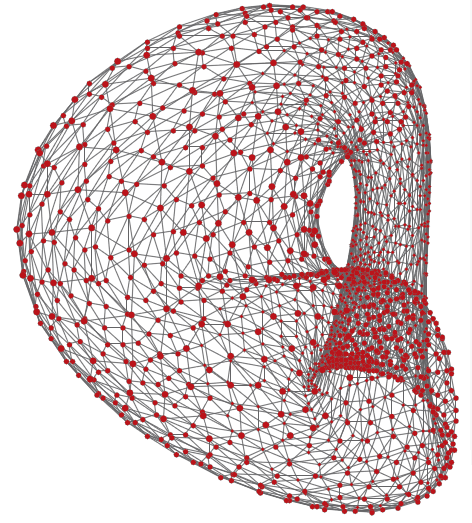}
  \hspace{.1\linewidth}
  \fbox{\includegraphics[width=.35\linewidth]{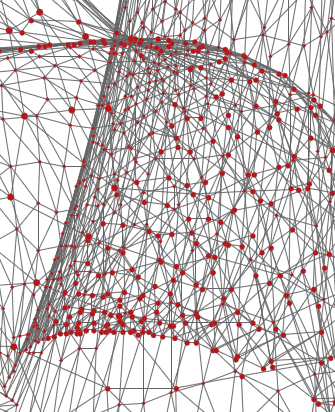}}
  \caption{\label{fig:klein}The Klein bottle is not a manifold in
  $\mathbb{R}^3$ (above): singular units in the SOAM mark the
  self-intersections. The same surface is a manifold in $\mathbb{R}^4$ and could
  be correctly reconstructed (below).}
\end{figure}
\subsection{When adaptation fails}

Two kinds of symptoms may signal the failure of the network adaptation process:
\begin{enumerate}
  \item[a)]Persistence of non-\emph{regular} units, typically \emph{connected}
  ones
  \item[b)]Persistence of \emph{singular} units
\end{enumerate}

The causes of a) could be twofold. Either the input dataset is not
a $\delta$-sample of the manifold $M$, in the sense of Theorem
\ref{thm:finiteSample}, or, in the light of the same theorem, the value of
$r_{min}$ is too low, in the sense that the network eventually became an
$\varepsilon$-sample too fine for the $\delta$-sample represented by the
input dataset.

Symptom b), on the other hand, could be due either to a too high value for
$r_{min}$, which can be fixed easily, or to noise. In agreement with Theorem
\ref{thm:finiteSample}, experiments show that the effects of noise on
convergence are `on-off': up to a certain value of the noise threshold, the
process, albeit slowed down, continues to converge; beyond that value, symptom
b) always occurs and the adaptation is doomed to fail. Given that the actual
noise threshold depends on the global feature size of $M$, its value may be
difficult to compute beforehand. As a compensation, SOAM failures are typically
very localized, thus showing which parts of $M$ are more problematic.

\section{Conclusions and future work}

The design of the SOAM algorithm relies heavily on the theoretical corpus
presented in the previous sections. Experiments show that these results
may be very effective in enforcing stronger topological guarantees for growing
neural networks. Clearly, many aspects still remain to be clarified. For one,
the guarantee that a SOAM will eventually reach a stable configuration - in
the sense described above - for any non-defective input dataset remains to be
proved.

Nevertheless, at least in the author's opinion, a strong point of the SOAM
algorithm is that it produces clear signs of a convergence - or lack thereof -
whose correctness, in the line of principle, is verifiable. In addition,
experiments seem to suggest that, in some sense, the algorithm exceeds known
theoretical results and it may lead to new solutions for known problems
\cite{Guibas-Oudot07}.

About future developments, the current limitations about the dimension of the
manifold might be overcome along the lines suggested in
\cite{Boissonnat-etal07}, although this would require switching to the
\emph{weighted} version of the Delaunay complex. Nonetheless, again in the
author's opinion, a great potential of the SOAM algorithm presented relates to
tracking non-stationary input manifolds, which is the reason why, in designing
the algorithm itself, care has been taken to keep all transitions reversible.

\section*{Acknowledgment}

The author would like to thank Niccol\`o Piarulli, Matteo Stori and many
others for their help with the experiments and support in making this work possible.


\end{document}